\newtheorem{theorem}{Theorem}
\newtheorem{proposition}{Proposition}

\documentclass[12pt]{article}

\usepackage{amsmath}
\usepackage{amssymb}
\usepackage{amsfonts}
\usepackage{graphicx}

\newcommand{\qed}{{\mbox{} \hspace*{\fill}{\vrule height5pt width4pt depth0pt}}\\}

\def\M{\hspace*{0.75em}}

\begin{document}

\title{Inventory Control Involving Unknown Demand of Discrete Nonperishable Items - Analysis of a Newsvendor-based Policy}
\author{Michael N. Katehakis, Jian Yang, and Tingting Zhou\\
Department of Management Science and information Systems\\
Business School, Rutgers University, Newark, NJ 07102}
\date{October 2015}
\maketitle

\begin{abstract}
Inventory control with unknown demand distribution is considered, with emphasis placed on the case involving discrete nonperishable items. We focus on an adaptive policy which in every period uses, as much as possible, the optimal newsvendor ordering quantity for the empirical distribution learned up to that period. The policy is assessed using the regret criterion, which measures the price paid for ambiguity on demand distribution over $T$ periods. When there are guarantees on the latter's separation from the critical newsvendor parameter $\beta=b/(h+b)$, a constant upper bound on regret can be found. Without any prior information on the demand distribution, we show that the regret does not grow faster than the rate $T^{1/2+\epsilon}$ for any $\epsilon>0$. In view of a known lower bound, this is almost the best one could hope for. Simulation studies involving this along with other policies are also conducted.

\vspace*{3mm} \noindent{\bf Keywords: }Inventory Control; Newsvendor; Critical Quantile; Empirical Distribution; Large Deviation; Information Theory
\end{abstract}

\newpage

\section{Introduction}

For a given firm, inventory control is about dynamically adjusting ordering quantities to minimize the total long-run expected cost. In traditional models, demand levels faced by a firm are often assumed to be random, however, with known probabilistic distributions. Even the knowledge on demand distribution can often prove to be too optimistic. When the firm has just introduced a new product or when its external environment has just transitioned to a previously unfamiliar phase (such as a severe economic downturn), it will not be sure of the demand patterns to be encountered. One way out is through the Bayesian approach. In it, the firm possesses a prior distribution on potential demand patterns. Then, posterior understanding on demand is updated by its realized levels. Inventory management taking this approach can be found in, for instance, Scarf \cite{S59} and Lariviere and Porteus \cite{LP99}.

Most other times, even a prior distribution on demand can seem far-fetched. What meager information one possesses might just be a collection of potential demand distributions. Now, the concerned firm has still to make decisions based on its past observations. But its goal is no longer about catering to a specific demand distribution or even a series of posterior demand distributions. Rather, its history-dependent (henceforward called adaptive) control policy should better yield results that are reasonably good under all potential demand distributions from the given collection. %In order to gauge the effectiveness of an adaptive policy, we adopt the regret criterion.
A given policy's regret under a given demand distribution and over a fixed time horizon measures the price paid for ambiguity; namely, it registers the difference between the policy's performance and that of the best policy tailor-made for the demand distribution had it been known. A policy will be considered good when its worst regret over all demand distributions in a collection grows as slowly as possible over time. %Naturally, the minimization of the worst regret over any given time period across all potential patterns in the given collection, or to a lesser degree, the asymptotic minimization of the worst regret as the length of the time period goes to infinity, poses as a prime target for researches on such adaptive policies.
%Often, this lofty goal is hard to attain. What can be done instead are bounds from two sides. On the one side, lower bounds may be found so that no adaptive policies can achieve regrets that are below them; on the other side, deftly crafted policies may offer upper-bounding guarantees on regrets. Of course, the convergence of these two types of bounds into one, a rare occurrence, would imply the optimality of the identified policies in terms of regret minimization.

In this paper we follow the  frequentist approach. This approach was  pioneered in the work of 
Robbins \cite{R52},  Lai and Robbins \cite{LR85},  Katehakis and Robbins \cite{Katehakis and Robbins 1995}, and  Burnetas and Katehakis \cite{BK96}, for  allocation problems,  where the main concern is on adaptively  selecting the most promising population to 
 draw from so as to maximize the total expected value of samples,   or equivalently to minimize the regret due to ignorance of the 
distributions.  
Later, the approach was brought to bear on adaptive Markov decision processes; see, e.g., Burnetas and Katehakis \cite{BK97}. For recent work in this area we refer to Burnetas et al \cite{BKK15}, and Cowan and Katehakis \cite{CK15gr}, \cite{CK15as}. 

Adaptive policies for inventory control have been considered. Huh and Rusmevichientong \cite{HR09} analyzed a gradient-based policy most suitable to the continuous-demand case. Their policy could also be thought of as an extension of stochastic approximation (SA), which started with Robbins and Monro \cite{RM51} and Kiefer and Wolfowitz \cite{KW52}. More recently, Besbes and Muharremoglu \cite{BM13} focused on the discrete-demand case of the repeated newsvendor problem and proposed policies with provably good performance guarantees.

In a revenue management setup, Besbes and Zeevi \cite{BZ09} studied the dynamic setting of prices while learning demand on the fly. Perakis and Roels \cite{PR08} minimized the worst regret for a single-period newsvendor problem. Rather than dynamic learning, they used conic optimization to deal with partial information on random demands in the form of known moments. For the newsvendor problem and its multi-period version involving nonperishable items, Levi, Roundy, and Shmoys \cite{LRS07} relied on randomly generated  demand samples to reach solutions with relatively good qualities at high probabilities. In this work, demand learning was through a black box capable of turning up an arbitrary number of samples at any time, rather than through the sequence of actually encountered demand levels.

We study inventory control involving the online learning of unknown demand, focusing on nonperishable items like Huh and Rusmevichientong \cite{HR09} and discrete-quantity analysis like Besbes and Muharremoglu \cite{BM13}. In many real-life situations ranging from manufacturing to retailing, nonperishability of items is an essential feature to be faced head on. Also, many applications, such as the management of bulk items, dictate that demand be discrete. An adaptation to Huh and Rusmevichientong's \cite{HR09} policy could work for the discrete-demand case, as demonstrated in the repeated-newsvendor analysis in their Section 3.4. However, further adaption seems needed for the nonperishable-inventory case; see~(\ref{lala}) and~(\ref{lalala}) later in our simulation study. %The regret analysis of this more complicated policy is so far unknown.
To the best of our knowledge, our theoretical performance guarantees on an adaptive inventory policy involving unknown demand of discrete nonperishable items have made original contributions.

We adopt a very simple and natural policy, the one that always orders, as much as possible, to the critical newsvendor quantile corresponding to the empirical demand distribution. The optimal ordering quantity for a newsvendor problem involving holding cost $h$, backlogging or effective lost sales cost $b$, and known demand distribution $f$, is the $\beta$-quantile of the distribution $f$, where $\beta=b/(h+b)$. By the beginning of period $t$, the empirical distribution $\hat f_{t-1}$ one has about demand is defined by the frequencies of various demand levels reached in periods $1,2,...,t-1$. The heuristic policy advocates ordering up to the $\beta$-quantile of $\hat f_{t-1}$ in every period $t$. It has been considered by Besbes and Muharremoglu \cite{BM13}. Our nonperishable-inventory variant needs to further ensure that items left over from earlier periods are accounted for. This is a difficult point that requires quite careful treatments.

%Our main contribution lies in the analysis of the newsvendor-based policy in the current inventory control setting involving unknown demand distributions. There are two cases.
We analyze two cases. In the first case in which the $\beta$-quantile estimation of none of the demand distributions considered is overly sensitive to small errors, we show that the worst regret will be bounded by a time-invariant constant.
%This result can be considered an extension of Besbes and Muharremoglu's \cite{BM13} Theorem 2, under their Assumption iii, to the current nonperishable-item case.
Though the flat rate over time is impressive, the result can be faulted by its requirement on the unknown distribution's behavior around its $\beta$-quantile. We thus go on to the second, more involved, case where all potential demand distributions are allowed. Given any $\epsilon>0$, we show that the worst regret over all distributions will not grow faster than the rate $T^{1/2+\epsilon}$. In view of the $T^{1/2}$-sized lower bound achieved at Lemma 4 of Besbes and Muharremoglu \cite{BM13}, this is almost the best one could hope for.
%Besides the perisability/nonperishability of items, the all-demand bound also sets this paper apart from the aforementioned earlier paper. In addition, o
Our derivation invokes large-deviation and information-theoretic results such as Sanov's Theorem and Pinsker's inequality, as well as innate features of empirical distributions and inventory control. Methodological advances might find applications elsewhere.

%lves novel uses of (a) , which gives a large-deviation bound on the difference between empirical distributions and their generators, (b) Pinsker's inequality, which bounds the total variation between two distributions by their Kullback-Liebler divergence, and (c) innate features of empirical distributions. These methodological advances might find applications elsewhere.

Our simulation study indicates the high likelihood with which the worst regret grows at the $T^{1/2}$-rate. Thus, it remains as a  future research item whether the $\epsilon>0$ in our upper bound can be removed. The study also shows that the newsvendor-based policy compares favorably with the adaptation to Huh and Rusmevichientong's \cite{HR09} SA-based policy. This is expected, as the former uses more information about past demand realizations. Through the study, we also confirm that the underlying distribution's separation from  $\beta$ plays a prominent role in determining the regret generated by the newsvendor-based policy.

The remainder of the paper is organized as follows.  Section~\ref{easy} introduces notation, problem formulation, and the newsvendor-based policy. The time-invariant bound with demand restriction around $\beta$ is given in Section~\ref{1bd}, while the slower-over-time bound without any demand restriction is derived in Section~\ref{2bd}. We present results of our simulation study in Section~\ref{computation}. The paper is concluded in Section~\ref{conclusions}.

\section{Problem and Policy}\label{easy}

We consider a multi-period inventory control problem in which unsatisfied demand is either backlogged or lost. Also, items are nonperishable so that those unsold in one period are carried over to the next period. Demand $D_{t}$ in each period $t=1,2,....$ is a random draw from a distribution with discrete support $\{0,1,...,\bar d\}$, where $\bar d$ is some positive integer. A generic distribution is representable by a vector $f\equiv (f(0),f(1),...,f(\bar d))$ in the $\bar d$-dimensional simplex within $\mathbb{R}^{\bar d+1}$:
\begin{equation}
\Delta\equiv\{f\in [0,1]^{\bar d+1}|\sum_{d=0}^{\bar d} f(d)=1\}.
\end{equation}
We use $F_f$ to denote the cumulative distribution function (cdf) associated with any given $f\in \Delta$. It satisfies $F_f(d)=\sum_{d'=0}^d f(d')$ for $d'=0,1,...,\bar d$.

We suppose production cost is linear at a unit rate $c$. %We also suppose inventory holding cost rate is $h$ per item per period and backlogging cost rate is $b$ per item per period.
The concerned planning horizon constitutes periods $1,2,...,T$. In the terminal period $T$, the firm will gain $cx$ if it has $x=1,2,...$ leftover items and will need to pay $-cx$ if $x=-1,-2,...$. Also, assume strictly positive holding cost rate $h$ and strictly positive backlogging or effective lost sales cost rate $b$. In the backlogging case, $b$ is usually in the same order of magnitude as $h$; whereas, in the lost sales case, $b$ is in the order of an item's profit margin and normally significantly greater than $h$. When the firm orders up to $y_t$ and experiences demand $d_t$ in each period $t=1,2,...,T$, its total cost in the backlogging case will be
\begin{equation}
c\cdot\sum_{t=1}^T d_t+\sum_{t=1}^T [h\cdot(y_t-d_t)^+ +b\cdot(d_t-y_t)^+].
\end{equation}
Since the first term in the above is not affected by the decision sequence $(y_1,y_2,...,y_T)$, we shall focus on the latter inventory-related cost term. In the lost sales case, the same conclusion can be reached when $b$ is an item's profit margin plus the actual per-item lost sales cost; see Huh and Rusmevichientong \cite{HR09}.

Given $f\in\Delta$, $t=1,2,...$, and real-valued function $g$ defined on $\{0,1,...,\bar d\}^t$, we use $\mathbb{E}_{f}[g(D_1,...,D_t)]$ to represent the average of $g(d_1,...,d_t)$ when each $D_s$ is independently sampled from distribution $f$:
\begin{equation}
\mathbb{E}_f[g(D_1,...,D_t)]\equiv\sum_{d_1=0}^{\bar d}\cdots\sum_{d_t=0}^{\bar d}f(d_1)\times\cdots\times f(d_t)\times g(d_1,...,d_t).
\end{equation}
For subset $D'\subset\{0,1,...,\bar d\}^t$, we understand $\mathbb{P}_f[D']$ by $\mathbb{E}_f[{\bf 1}((D_1,...,D_t)\in D')]$. Note that
\begin{equation}
\mathbb{P}_f[D_t\leq d]=\mathbb{E}_f[{\bf 1}(D_t\leq d)]=\sum_{d'=0}^{d}f(d')=F_f(d).
\end{equation}

Define $Q_f(y)$ for every $f\in\Delta$ and $y=0,1,...,\bar d$, so that
\begin{equation}\label{def0}\begin{array}{ll}
Q_f(y)&\equiv\mathbb{E}_f[h\cdot (y-D)^+ + b\cdot(D-y)^+]=\sum_{d=0}^{\bar d} f(d)\cdot(h\cdot(y-d)^++b\cdot(d-y)^+)\\
&=h\cdot\sum_{d=0}^{y-1}F_f(d)+b\cdot\sum_{d=y}^{\bar d-1}(1-F_f(d)).
\end{array}\end{equation}
It is the single-period average cost under order-up-to level $y$. Let $Q^*_f=\min_{y=0,1,...,\bar d}Q_f(y)$ be the minimum cost in one period under $f$. Suppose $y^*_f$ is an order-up-to level that achieves the one-period minimum. Then, when facing a $T$-period horizon, an optimal policy will be to repeatedly order up to this level. Thus, the minimum cost over $T$ periods is $Q^*_f\cdot T$.

A salient feature of our current problem, however, is that $f$ is not known beforehand. So instead of any $f$-dependent policy, we seek a good $f$-independent policy which takes advantage of demand levels observed in the past. An adaptive policy ${\bf y}\equiv (y_1,y_2,...)$ is such that, for $t=1,2,...$, each $y_t=0,1,...,\bar d$ is a function of the historical demand vector ${\bf d}_{[1,t-1]}\equiv (d_1,...,d_{t-1})\in \{0,1,...,\bar d\}^{t-1}$. Under it, the $T$-period total average cost is
\begin{equation}
Q^T_f({\bf y})\equiv\sum_{t=1}^T \mathbb{E}_f[h\cdot (y_t({\bf D}_{[1,t-1]})-D_t)^+ +b\cdot (D_t-y_t({\bf D}_{[1,t-1]}))^+]=\sum_{t=1}^T \mathbb{E}_f[Q_f(y_t({\bf D}_{[1,t-1]}))],
\end{equation}
where the second equality comes from the independence between ${\bf D}_{[1,t-1]}$ and $D_t$. Now define $T$-period regret $R^T_f({\bf y})$ of using the adaptive policy ${\bf y}$ against the unknown distribution $f$:
\begin{equation}\label{r-def}
R^T_f({\bf y})\equiv Q^T_f({\bf y})-Q^*_f\cdot T=\sum_{t=1}^T \mathbb{E}_f[Q_f(y_t({\bf D}_{[1,t-1]}))]-Q^*_f\cdot T.
\end{equation}
Here, the ultimate goal should be that of identifying adaptive policies ${\bf y}$ that prevent $R^T_f({\bf y})$ from growing too fast in $T$ under all or at least most $f$'s within $\Delta$.

We concentrate on one policy inspired by an optimal $y^*_f$ when $f$ is known. From~(\ref{def0}), we see that necessary and also sufficient conditions for optimality of any $y$ are
\begin{equation}\label{cond1}
Q_f(y+1)-Q_f(y)= %h\cdot\mathbb{P}_f[D\leq y]-b\cdot\mathbb{P}_f[D\geq y+1]=
(h+b)\cdot F_f(y)-b\geq 0,
\end{equation}
and
\begin{equation}\label{cond2}
Q_f(y)-Q_f(y-1)= %h\cdot\mathbb{P}_f[D\leq y-1]-b\cdot\mathbb{P}_f[D\geq y]=
(h+b)\cdot F_f(y-1)-b\leq 0,
\end{equation}
%In other words, we are supposed to have
%\begin{equation}
%\frac{\mathbb{P}_{f^*}[D\leq y^*-1]}{\mathbb{P}_{f^*}[D\geq y^*]}\leq \frac{b}{h}\leq\frac{\mathbb{P}_{f^*}[D\leq y^*]}{\mathbb{P}_{f^*}[D\geq y^*+1]}.
%\end{equation}
Let $\beta=b/(h+b)$ be the famous newsvendor parameter that lies in $(0,1)$. For $f\in\Delta$, let $y^*_{f,\beta}$ be the associated newsvendor order-up-to level, so that
\begin{equation}\label{yf-def}
y^*_{f,\beta}\equiv F_f^{\;-1}(\beta)\equiv\min\{d=0,1,...,\bar d|F_f(d)\geq \beta\}.
\end{equation}
By definition, $F_f(y^*_{f,\beta})\geq \beta$ and hence $Q_f(y^*_{f,\beta}+1)-Q_f(y^*_{f,\beta})\geq 0$ by~(\ref{cond1}); also, $F_f(y^*_{f,\beta}-1)<\beta$ and hence $Q_f(y^*_{f,\beta})-Q_f(y^*_{f,\beta}-1)<0$ by~(\ref{cond2}). Therefore, $Q_f(y^*_{f,\beta})=Q^*_f$, meaning that $y^*_{f,\beta}$ is an optimal order-up-to level for the one-period problem when $f$ is known. %We are to take advantage of this face.
Now with $f$ unknown, we might adopt level $y^*_{f_{t-1},\beta}$ where $f_{t-1}$ is a good estimate of $f$ after observing demand vector  $D_{[1,t-1]}$. .

The prime candidate for $f_{t-1}$ is the empirical distribution $\hat f_{t-1}$. For $t=2,3,...$, define $\hat f_{t-1}\in\Delta$ by $(\hat f_{t-1}(0),\hat f_{t-1}(1),...,\hat f_{t-1}(\bar d))$, so that for every $d=0,1,...,\bar d$,
\begin{equation}\label{emp}
 \hat f_{t-1}(d)=\frac{\sum_{s=1}^{t-1} {\bf 1}(d_s=d)}{t-1}.
\end{equation}
Each $\hat f_{t-1}$ has its corresponding cdf $\hat F_{t-1}\equiv F_{\hat f_{t-1}}$. Both $\hat f_{t-1}$ and $\hat F_{t-1}$ are certainly functions of the past demand vector $d_{[1,t-1]}\equiv (d_1,d_2,...,d_{t-1})$. However, for notational simplicity we have refrained from making this dependence explicit.
%Both are dependent on past-demand vector ${\bf d}_{[1t]}$.
Our heuristic policy applies the newsvendor formula to the empirical demand distribution. It lets the firm order nothing in period 1; that is, $y_1=\hat y_1=0$. For any $t=2,3,...$, it advises the firm to order up to
\begin{equation}\label{yt-def}
y_t=\hat y_t\vee (y_{t-1}-d_{t-1}),
\end{equation}
in period $t$, in which
\begin{equation}\label{newsboy}
\hat y_t=y^*_{\hat f_{t-1},\beta}=\hat F_{t-1}^{\;-1}(\beta)\equiv F_{\hat f_{t-1}}^{\;-1}(\beta)=\min\{d=0,1,...,\bar d|\sum_{s=1}^{t-1}{\bf 1}(d_s\geq d)\geq \beta\cdot (t-1)\}.
\end{equation}

For simplicity, we have not made the dependence of $y_t$ and $\hat y_t$ on $d_{[1,t-1]}$ explicit. Nor have we used the full $d_{[1,t-2]}$-dependent notation on $y_{t-1}$. This will apply to the remainder of the paper. For the lost sales case, we need to guarantee that $y_t\geq 0$ and hence enhance~(\ref{yt-def}) to $y_t=\hat y_t\vee (y_{t-1}-d_{t-1})^+$. However, the current heuristic through~(\ref{newsboy}) has ensured that $\hat y_t\geq 0$. So the same~(\ref{yt-def}) can still be used. Of course, a typical $\beta$ for the backlogging case might be around $1/2$, whereas a typical $\beta$ for the lost sales case might be close to 1.

Our main purpose is to show that the $f$-blind and yet adaptive policy ${\bf y}$ described by~(\ref{yt-def}) and~(\ref{newsboy}) will incur regret $R^T_f({\bf y})$ as defined by~(\ref{r-def}) that is slow-growing in the planning length $T$ for most or even all $f$'s among $\Delta$. The requirement of $y_t\geq y_{t-1}-d_{t-1}$ in~(\ref{yt-def}), as necessitated by our nonperishable-inventory setting, renders decisions made over different periods more entangled with one another. Along with the discrete-demand setup, this substantially complicates the problem's analysis. %Not surprisingly, some of the proofs, especially that for Proposition~\ref{2-bound-hp}, are quite involved.

\section{Separation-affected Bound}\label{1bd}

We first establish a bound for $R^T_f({\bf y})$ when there are guarantees on the distances between the $F_f(d)$ values and the critical value $\beta$. By~(\ref{r-def}) and~(\ref{yt-def}),
\begin{equation}\label{regret0}
R^T_f({\bf y})=R^{T1}_f({\bf y})+R^{T2}_f({\bf y}),
\end{equation}
where
\begin{equation}\label{def1}
R^{T1}_f({\bf y})=\sum_{t=1}^T \mathbb{E}_f[Q_f(\hat{y}_t)]-Q^*_f\cdot T,
\end{equation}
and, since $y_1=\hat y_1=0$ by design and hence $y_2=\hat y_2$,
\begin{equation}\label{def2}
R^{T2}_f({\bf y})=\sum_{t=3}^{T}\mathbb{E}_f[Q_f(y_t)-Q_f(\hat{y}_t)].
\end{equation}
It might be said that $R^{T1}_f({\bf y})$ represents the price paid for the regrettable fact that the policy ${\bf y}$ was not designed with the particular distribution $f$ in mind; meanwhile, $R^{T2}_f({\bf y})$ captures the additional cost due to the nonperishability of items. We find it convenient to bound $R^{T1}_f({\bf y})$ and $R^{T2}_f({\bf y})$ separately.

Due to $y^*_{f,\beta}$'s definition in~(\ref{yf-def}), we must have $f(y^*_{f,\beta})>0$; for otherwise, $y^*_{f,\beta}$ could be made even smaller. Now there are two cases, with

\indent\M Case 1: $\alpha_{f,\beta}\equiv F_f(y^*_{f,\beta}-1)<\beta<\gamma_{f,\beta}\equiv F_f(y^*_{f,\beta})$; and,

\indent\M Case 2: $\alpha_{f,\beta}\equiv F_f(y^*_{f,\beta}-1)<\beta=F_f(y^*_{f,\beta})$.

We concentrate on case 1 first. Let us use $x\in [0,1]$ to denote the Bernoulli distribution where the chance for 1 is $x$ and that for 0 is $1-x$. Then, the numbers $\alpha_{f,\beta}=F_f(y^*_{f,\beta}-1)$ and $\gamma_{f,\beta}=F_f(y^*_{f,\beta})$ represent two Bernoulli distributions. Use $\hat \alpha_{t-1}$ and $\hat\gamma_{t-1}$ for the empirical distributions of $\alpha_{f,\beta}$ and $\gamma_{f,\beta}$, respectively, both recording frequencies of 1's in the first $t-1$ Bernoulli draws. One important observation is that
\begin{equation}\label{ineq1}\left\{\begin{array}{ll}
\mathbb{P}_f[\hat F_{t-1}^{\;-1}(\beta)\leq y^*_{f,\beta}-1]=\mathbb{P}_f[\hat F_{t-1}(y^*_{f,\beta}-1)\geq \beta]=\mathbb{P}_{\alpha_{f,\beta}}[\hat \alpha_{t-1}\geq \beta],\\
\mathbb{P}_f[\hat F_{t-1}^{\;-1}(\beta)\geq y^*_{f,\beta}+1]=\mathbb{P}_f[\hat F_{t-1}(y^*_{f,\beta})<\beta]=\mathbb{P}_{\gamma_{f,\beta}}[\hat \gamma_{t-1}<\beta].
\end{array}\right.\end{equation}
%In the current case 1,
%\begin{equation}\label{delta-def}
%\delta_f\equiv\min\{\beta-\alpha_{f,\beta},\gamma_{f,\beta}-\beta\}>0.
%\end{equation}

By a special version of Sanov's Theorem (Dembo and Zeitouni \cite{DZ98}, (2.1.12)), we have upper bounds for right-hand sides above:
\begin{equation}\left\{\label{ineq2}\begin{array}{l}
\mathbb{P}_{\alpha_{f,\beta}}[\hat \alpha_{t-1}\geq \beta]\leq t^2\cdot\exp(-(t-1)\cdot\inf_{x\in [\beta,1]}\mbox{D}_{KL}(x||\alpha_{f,\beta})),\\
\mathbb{P}_{\gamma_{f,\beta}}[\hat \gamma_{t-1}<\beta]\leq t^2\cdot\exp(-(t-1)\cdot\inf_{x\in [0,\beta)}\mbox{D}_{KL}(x||\gamma_{f,\beta})),
\end{array}\right.\end{equation}
respectively, where
\begin{equation}
\mbox{D}_{KL}(u||v)\equiv u\cdot\ln(\frac{u}{v})+(1-u)\cdot\ln(\frac{1-u}{1-v}), \end{equation}
i.e., the relative entropy or Kullback-Leibler divergence between Bernoulli distributions $u$ and $v$. Since $\mbox{D}_{KL}(\cdot||v)$ is known to be convex with minimum achieved at $v$ (Cover and Thomas \cite{CT06}, Theorems 2.6.3 and 2.7.2),
\begin{equation}\label{ineq3}\left\{\begin{array}{l}
\inf_{x\in [\beta,1]}\mbox{D}_{KL}(x||\alpha_{f,\beta})=\mbox{D}_{KL}(\beta||\alpha_{f,\beta})>0,\\
\inf_{x\in [0,\beta)}\mbox{D}_{KL}(x||\gamma_{f,\beta})=\mbox{D}_{KL}(\beta||\gamma_{f,\beta})>0.
\end{array}\right.\end{equation}
Now
\begin{equation}\label{kappa-def}
\kappa_{f,\beta}\equiv\min\{\mbox{D}_{KL}(\beta||\alpha_{f,\beta}),\mbox{D}_{KL}(\beta||\gamma_{f,\beta})\}>0.
\end{equation}

Combining~(\ref{ineq1}) to~(\ref{kappa-def}), we have both $\mathbb{P}_f[\hat F_{t-1}^{\;-1}(\beta)\leq y^*_{f,\beta}-1]$ and $\mathbb{P}_f[\hat F_{t-1}^{\;-1}(\beta)\geq y^*_{f,\beta}+1]$ being bounded from above by $t^2\cdot\exp(-\kappa_{f,\beta}\cdot (t-1))$. In view of~(\ref{newsboy}),
\begin{equation}\label{useful1}
\mathbb{P}_f[\hat y_t\leq y^*_{f,\beta}-1]\vee\mathbb{P}_f[\hat y_t\geq y^*_{f,\beta}+1]\leq \min\{t^2\cdot\exp(-\kappa_{f,\beta}\cdot(t-1)),1\}.
\end{equation}
We can identify a large enough $\tau_{f,\beta}$, so that for $t\geq \tau_{f,\beta}+1$, both
\begin{equation}\label{aha2}
t^{2}\cdot \exp(-\kappa_{f,\beta}\cdot (t-1))<\frac{1}{2},
\end{equation}
and
\begin{equation}\label{aha1}
\frac{(t+1)^{2}\cdot \exp(-\kappa_{f,\beta}\cdot t)}{t^{2}\cdot \exp(-\kappa_{f,\beta}\cdot (t-1))}<\exp(-\frac{\kappa_{f,\beta}}{2}).
\end{equation}

Now $R^{T1}_f({\bf y})$ as given in~(\ref{def1}) can be bounded.

\begin{proposition}\label{1-bound}
Under case 1, it is true that
\[R^{T1}_f({\bf y})\leq (h\bar d+b\bar d)\cdot (\tau_{f,\beta}+\frac{1}{2\cdot(1-\exp(-\kappa_{f,\beta}/2))}).\]
\end{proposition}

All proofs of this section can be found in Appendix~\ref{appendix1}. Proposition~\ref{1-bound} demonstrates that there is a $T$-independent bound for $R^{T1}_f({\bf y})$; however, the constant is heavily dependent on the relative positioning between the unknown distribution $f$ and the parameter $\beta$. It conveys the same message as Besbes and Muharremoglu's \cite{BM13} Theorem 2 on the repeated newsvendor problem. For our case involving nonperishable items, we still need a bound on $R^{T2}_f({\bf y})$ defined in~(\ref{def2}). For this purpose, suppose $\epsilon_f\equiv f(\bar d)>0$.

\begin{proposition}\label{2-bound}
Under case 1, it is true that
\[R^{T2}_f({\bf y})\leq h\bar d\cdot (\tau_{f,\beta}+\frac{1}{1-\exp(-\kappa_{f,\beta}/2)}+\frac{1-\epsilon_f}{\epsilon_f}+\frac{1}{2\epsilon_f\cdot (1-\exp(-\kappa_{f,\beta}/2))}).\]
\end{proposition}

The proof revolves around bounding $\mathbb{P}_f[y_{t+1}\geq \hat y_{t+1}+1]$ for $t$ large enough. When case 2 occurs, let $y^1_{f,\beta}\geq y^*_{f,\beta}$ be such that
\begin{equation}\label{posi} \beta=F_f(y^*_{f,\beta})=F_f(y^*_{f,\beta}+1)=\cdots=F_f(y^1_{f,\beta})<F_f(y^1_{f,\beta}+1).\end{equation}
From~(\ref{cond1}), we see that
\begin{equation} Q_f(y^*_{f,\beta})=Q_f(y^*_{f,\beta}+1)=\cdots=Q_f(y^1_{f,\beta}+1)<Q_f(y^1_{f,\beta}+2).\end{equation}
So instead of $\mathbb{P}_f[\hat y_t\geq y^*_{f,\beta}+1]$, we need only to estimate $\mathbb{P}_f[\hat y_t\geq y^1_{f,\beta}+2]$. But according to~(\ref{ineq1}), this is the same as $\mathbb{P}_{\gamma_{f,\beta}}[\hat\gamma_{t-1}<\beta]$ for $\gamma_{f,\beta}=F_f(y^1_{f,\beta}+1)$. Now due to~(\ref{posi}),  $\min\{\beta-\alpha_{f,\beta},\gamma_{f,\beta}-\beta\}=\min\{f(y^*_{f,\beta}),f(y^1_{f,\beta}+1)\}>0$. So the same bounds as in Propositions~\ref{1-bound} and~\ref{2-bound} can be established.
A definition suitable for both cases is that
\begin{equation}\label{ag-def}
\alpha_{f,\beta}\equiv\max\{F_f(d)|F_f(d)<\beta\},\hspace*{.8in}\gamma_{f,\beta}\equiv\min\{F_f(d)|F_f(d)>\beta\}.
\end{equation}

Now we can achieve an upper bound for $R^T_f({\bf y})$, regardless of the case we are in. For any $f\in\Delta$, define $f$'s separation from $\beta$ by
\begin{equation}\label{small-def}
\delta_{f,\beta}\equiv (\beta-\alpha_{f,\beta})\wedge (\gamma_{f,\beta}-\beta).
\end{equation}
Given $\epsilon,\delta>0$, define $\Delta_{\epsilon,\delta,\beta}\subset\Delta$ to be the collection of $f$'s with guaranteed lower bounds on both $\epsilon_f$ and $\delta_{f,\beta}$:
\begin{equation}\label{delta-def}
\Delta_{\epsilon,\delta,\beta}\equiv\{f\in\Delta|\epsilon_f\geq \epsilon,\delta_{f,\beta}\geq \delta\}.
\end{equation}
It turns out that $R^T_f({\bf y})$ has an upper bound that is uniform across $f\in \Delta_{\epsilon,\delta,\beta}$.

\begin{theorem}\label{newsboy-up}
For any $\epsilon,\delta>0$, there is a positive constant $A_{\epsilon,\delta,\beta}$ so that
\[ \sup_{f\in \Delta_{\epsilon,\delta,\beta}}R^{T}_f({\bf y})\leq A_{\epsilon,\delta,\beta}.\]
\end{theorem}

Theorem~\ref{newsboy-up} shows that the regret $R^T_f({\bf y})$ is bounded from above by a constant independent of time $T$, so long as there are known lower bounds on both $\epsilon_f\equiv f(\bar d)$ and $\delta_{f,\beta}$, the separation between $f$ and $\beta$. Between the two requirements, the first one appears more reasonable as $\bar d$ can always be the highest level that demand can ever reach. The second requirement, on the other hand, straddles between both demand distributions and cost parameters. It seems far-fetched to exclude a priori distributions $f$ satisfying $\delta_{f,\beta}\in (0,\delta)$ from consideration.

\section{Universal Bound}\label{2bd}

Due to the shortcoming inherent in the previous bound, we feel compelled to derive a bound on $R^T_f({\bf y})$ that requires no prior knowledge on $f$, let alone its relative positioning with respect to the cost parameter $\beta$. The new analysis involves Sanov's Theorem which offers a large-deviational bound on the difference between an empirical distribution and its generating distribution, Pinsker's inequality which connects two distances between distributions, Markov's inequality, and innate properties of the inventory management problem.

Let $\mbox{D}_{KL}(g||f)$ be the relative entropy or Kullback-Leibler divergence between distributions $g$ and $f$ in $\Delta$, so that
\begin{equation}
\mbox{D}_{KL}(g||f)\equiv\sum_{d=0}^{\bar d}g(d)\cdot\ln(\frac{g(d)}{f(d)}) .\end{equation}
Sanov's Theorem (Dembo and Zeitouni \cite{DZ98}, (2.1.12)) states that, for any set $G$ of the demand space $\Delta$ that is closed in the Euclidean metric,
\begin{equation}\label{ssnv}
\mathbb{P}_f[\hat f_{t-1}\in G]\leq t^{\bar d+1}\cdot\exp(-(t-1)\cdot\inf_{g\in G}\mbox{D}_{KL}(g||f)).
\end{equation}
Let $\delta_V(f,g)$ be the total variation between distributions $f$ and $g$; i.e.,
\begin{equation}
\delta_V(f,g)\equiv\max_{K\subset\{0,1,...,\bar d\}}\mid \sum_{d\in K}f(d)-\sum_{d\in K}g(d)\mid,
\end{equation}
which also equals $\mid\mid f-g\mid\mid_1/2\equiv\sum_{d=0}^{\bar d}\mid f(d)-g(d)\mid/2$. Pinsker's inequality (Cover and Thomas \cite{CT06}, Lemma 11.6.1) specifies that
\begin{equation}\label{pinsker}
\delta_V(f,g)\leq \sqrt{\frac{\mbox{D}_{KL}(g||f)}{2}}.
\end{equation}

With the absence of any prior knowledge on the $f\in\Delta$, we can manage to obtain a $T^{1/2+\epsilon}$-bound on the $R^{T1}_f({\bf y})$ defined in~(\ref{def1}) for any $\epsilon>0$. Due to~(\ref{newsboy}), the key is to show that $Q_f(y^*_{\hat f_{t-1},\beta})-Q_f(y^*_{f,\beta})$ will converge to 0 quickly. In view of the Sanov property~(\ref{ssnv}), this will be achievable if we can show that $Q_f(y^*_{g,\beta})-Q_f(y^*_{f,\beta})$ will be small when $g$ and $f$ are close by. This is when Pinsker's inequality~(\ref{pinsker}) and other properties related to the inventory management problem, such as the optimality of $y^*_{f,\beta}$ to $Q_f(\cdot)$ and the linearity of $Q_f(y)-Q_g(y)$ in the distance between $f$ and $g$, will be useful. The final form of the bound comes from the estimation of certain summations through integrations.

\begin{proposition}\label{1-bound-hp}
For any $\epsilon>0$, there are positive constants $A_\epsilon$ and  $B_\epsilon$, so that
\[ R^{T1}_f({\bf y})\leq A_\epsilon+B_\epsilon\cdot T^{1/2+\epsilon}.\]
\end{proposition}

All proofs of this section can be found in Appendix~\ref{appendix2}. The current result is ultimately about bounding the two sums  $\sum_{t=1}^T\sqrt{\varepsilon_t}$ and $\sum_{t=1}^T t^{\bar d+1}\cdot \exp(-\varepsilon_t\cdot (t-1))$ simultaneously for a sequence $\varepsilon_t$; see~(\ref{abv}) in our proof. The $T^{1/2+\epsilon}$-sized rate comes from our choice of $\varepsilon_t=(t-1)^{-1+2\epsilon}$ to balance the two terms. If we remove $\epsilon>0$, the first sum will give a $T^{1/2}$-sized rate but the second sum will fail to achieve a below-$T$ rate. This is why we can attain the $T^{1/2+\epsilon}$-sized rate for an arbitrarily small $\epsilon>0$ but never the exact $T^{1/2}$-sized rate.

Next, we obtain a bound in the same order of magnitude for $R^{T2}_f({\bf y})$ as defined by~(\ref{def2}). Let us first get some more understanding on the entity. From~(\ref{yt-def}), we see that
\begin{equation} \label{expp}
y_t=\hat y_t\vee (\hat y_{t-1}-d_{t-1})\vee(\hat y_{t-2}-d_{t-2}-d_{t-1})\vee\cdots \vee (\hat y_1-d_1-d_2-\cdots-d_{t-1}).
\end{equation}
There is a latest $s$ so that 
\begin{equation}\label{great}
y_t=\hat y_s-d_s-d_{s+1}-\cdots-d_{t-1},
\end{equation}
which occurs exactly when 
\begin{equation}\label{critical}
\hat y_s-d_s-d_{s+1}-\cdots-d_{t-1}-1\geq \hat y_t\vee (\hat  y_{t-1}-d_{t-1})\vee\cdots\vee (\hat y_{s+1}-d_{s+1}-\cdots-d_{t-1}),
\end{equation}
and
\begin{equation}\label{reit}
\hat y_s\geq (\hat y_{s-1}-d_{s-1})\vee(\hat y_{s-2}-d_{s-2}-d_{s-1})\vee \cdots \vee (\hat y_1-d_1-d_2-\cdots-d_{s-1}).
\end{equation}

Inspired by the above, we define random variables $I\geq 1$ and $S_1,S_2,...,S_I,S_{I+1}$ in an iterative fashion as follows. First, let $S_1=1$. Now for some $i=1,2,...$, suppose $S_i$ has been settled. Then, let $S_{i+1}$ be the first $t$ after $S_i$ so that
\begin{equation}\label{notso}
\hat y_t\geq \hat y_{S_i}-D_{S_i}-D_{S_i+1}-\cdots-D_{t-1},
\end{equation}
if such a $t\leq T$ can be identified. If not, mark the latest $i$ as $I$ and let $S_{I+1}=T+1$. For any $t$, let $L(t)$ be the largest $S_i\leq t$. This $L(t)$ can serve as the earlier $s$ satisfying~(\ref{critical}) and~(\ref{reit}) that corresponds to $t$. Note that $L(t)$ along with $D_{L(t)},D_{L(t)+1},...,D_{t-1}$ are independent of $D_t$.
% A key feature is that, if $L(t)\leq t-1$,
%\begin{equation}
%L(t)=L(t-1)=\cdots=L(L(t)-1)=L(L(t)).
%\end{equation}
So by~(\ref{def2}), as well as~(\ref{great}) to~(\ref{notso}),
\begin{equation}\label{ee74}
R^{T2}_f({\bf y})=\sum_{t=3}^T\mathbb{E}_f[Q_f(\hat y_{L(t)}-D_{L(t)}-\cdots-D_{t-1})-Q_f(\hat y_t)].
\end{equation}
Now we are in a position to derive the bound.

\begin{proposition}\label{2-bound-hp}
For any $\epsilon>0$, there exist positive constants $C_{\epsilon}$ and $D_\epsilon$ so that
\[ R^{T2}_f({\bf y})\leq C_{\epsilon}+D_{\epsilon}\cdot T^{1/2+\epsilon}.\]
\end{proposition}

This is the most difficult result of the paper. We will exploit the observations made from~(\ref{expp}) to~(\ref{ee74}) to the fullest extent, with the basic understanding that the actual order-up-to level $y_t$ will be $\hat y_s-D_s-\cdots-D_{t-1}$ for some $s\leq t$. We are tasked to show that the term $Q_f(\hat y_s-D_s-\cdots-D_{t-1})-Q_f(\hat y_t)$ can be bounded. For $\gamma=1-f(0)$, we divide the proof into two cases, the one with $\gamma\geq (1-\beta)/2$ and the other one with $\gamma<(1-\beta)/2$.

In the former large-$\gamma$ case, demand will accumulate over time with a guaranteed speed and $\hat y_t\geq \hat y_s-D_{s}-\cdots-D_{t-1}$ will occur ever more surely as $t-s$ increases. Then, for the minority case where $t-s$ is small, by exploiting natures of the empirical distribution and the newsvendor formula, we can come up with bounds related to $|Q_f(\hat y_s-D_s-\cdots-D_{t-1})-Q_f(\hat y_t)|\cdot{\bf 1}(\hat y_t\leq \hat y_s-D_{s}-\cdots-D_{t-1}-1)$. %However, the bound is $\gamma$-dependent and will blow if we let $\gamma$ shrink to zero.
Especially important is the observation that
$\hat y_t\leq\hat y_{s}-D_{s}-\cdots-D_{t-1}-1$ only if
\begin{equation}\label{dada-o}
\beta \leq\hat F_{t-1}(\hat y_t)
\leq \hat F_{t-1}(\hat y_{s}-D_{s}-\cdots-D_{t-1}-1)<\beta+\frac{t-s}{s};
\end{equation}
see~(\ref{dada}) later. We will end up with a trade-off already encountered in the proof of Proposition~\ref{1-bound-hp}; see~(\ref{chicago}) in the proof. This is the source of the $T^{1/2+\epsilon}$-sized growth rate. However, the constants will grow as $\gamma$ shrinks, because it takes ever longer for demand to accumulate.

Therefore, we seek a different approach for the latter small-$\gamma$ case, when $\gamma<(1-\beta)/2<1-\beta$. This is the time when $y^*_{f,\beta}=\hat F_f^{\;-1}(\beta)=0$ because $F_f(0)=f(0)=1-\gamma>\beta$. We utilize the fact that $\hat y_s-D_s-\cdots-D_{t-1}\geq 1$ is the bare minimum for $\hat y_s-D_s-\cdots-D_{t-1}\geq \hat y_t+1$. But the latter will be true only if both $\hat y_s\geq 1$ and for some $d=1,2,...,\bar d$,  both $\hat y_s\geq d$ and $D_s+\cdots+D_{t-1}\leq d-1$. For all $\gamma$'s in the interval $[0,(1-\beta)/2)$, we achieve a uniform bound in the order of $\ln T$, which is dominated by the one obtained in the first case.

Besides Sanov's Theorem and Pinsker's inequality, the proof also exploits Markov's inequality in bounding $\mathbb{P}_f[\hat y_s\geq d]$, which, through~(\ref{yf-def}) to~(\ref{newsboy}), is the chance for the portion of earlier demand levels at or exceeding $d$ to be greater than $1-\beta$. Combining Propositions~\ref{1-bound-hp} and~\ref{2-bound-hp}, %while realizing that $C_{\epsilon,\gamma}$ in the latter can be used for cases with $1-f(0)\in [\gamma,1]$,
we get a bound for $R^T_f({\bf y})$ that is not tangled up with how $f$ positions with $\beta$. %Let $\Delta_\gamma=\{f\in\Delta|f(0)\in [0,1-\gamma]\}$.

\begin{theorem}\label{newsboy-up-hp}
For any $\epsilon>0$, there are positive constants $E_{\epsilon}$ and $F_\epsilon$ so that
\[ \sup_{f\in \Delta} R^T_f({\bf y})\leq E_{\epsilon}+F_\epsilon\cdot T^{1/2+\epsilon}.\]
\end{theorem}

%The dominant term at large $T$ values is $T^{1/2+\epsilon}$.
The constants involved can depend on the problem's parameters $h$, $b$, and $\bar d$. However, they are uniform across all $f$'s in $\Delta$.
%\section{Lower Bound}
For the repeated newsvendor problem, Besbes and Muharremoglu \cite{BM13} has already shown a $T^{1/2}$-sized lower bound (Lemma 4, with $\varepsilon$ replaced by $1/\sqrt{T}$ in its (C-8)). %As probably already hinted to at Theorem~\ref{newsboy-up}, t
The example used for the bound involves distributions $f$ with small separations from $\beta$ but in opposite directions. According to~(\ref{yt-def}), the current case merely adds the restriction $y_{t}({\bf d}_{[1,t-1]})\geq y_{t-1}({\bf d}_{[1,t-2]})-d_{t-1}$ to the adaptive policy considered. So the lower bound can be no better.

In view of this, the above is almost the best one can hope for. Huh and Rusmevichientong's \cite{HR09} SA-based policy was shown to have a $T^{1/2}$-sized bound when ordering quantities are discrete. But the analysis was done for the repeated-newsvendor setting. The policy's adaptation to the nonperishable-item setting, as to occur in~(\ref{lala}) and~(\ref{lalala}), appears to be more complicated. Its full analysis awaits further research.

\section{Simulation Study}\label{computation}

In the study, we fix $h+b=10$ and $\bar d=20$. We let $b$'s variation drive changes in $\beta=b/(h+b)$. %Try $b=1$, $3$, $5$, $7$, $9$ and hence $\beta=0.1$, $0.3$, $0.5$, $0.7$, $0.9$.
Smaller $\beta$ values are suitable for the backlogging case and larger $\beta$ values the lost sales case. At each parameter combination, we randomly generate $K=1,000$ distributions $f$ in $\Delta$ uniformly. Within each run $k=1,2,...,K$, particularly, let $\xi_1,...,\xi_{\bar d}$ be independent random variables uniformly distributed on $[0,1]$. Rank them so that
\begin{equation}\label{replacable}
\eta_1\equiv\xi_{(1)}<\eta_2\equiv\xi_{(2)}<\cdots<\eta_{\bar d}\equiv \xi_{(\bar d)}.
\end{equation}
Also, let $\eta_0=0$ and $\eta_{\bar d+1}=1$. We create $f=(f(0),f(1),...,f(\bar d))$ so that
\begin{equation}
f(i)=\eta_{i+1}-\eta_i,\hspace*{.8in}\forall i=0,1,...,\bar d.
\end{equation}

Under each $f$ out of the $K$ possibilities, we test policies $a$ for $T=10,000$ periods on $L=100$ sample paths of demand levels. In each period $t$ on the $l$th path, we generate a random variable $\xi_{\bar d+(l-1)T+t}$ uniformly distributed on $[0,1]$. Then, let demand $d_t$ in that period be the only $d=0,1,...,\bar d$ that satisfies
\begin{equation}
\eta_{d-1}\equiv\sum_{i=0}^{d-1}f(i)\leq \xi_{\bar d+(l-1)T+t}<\sum_{i=0}^d f(i)\equiv \eta_d.
\end{equation}
Let $y^a_t$ be the order-up-to level in period $t$ under policy $a$. We use the following as an approximation to the policy's expected regret by time $t$:
\begin{equation}
r^{a,t}_{f,\beta}=\mbox{AVG}\{\sum_{s=1}^t [h\cdot (y^a_s-d_s)^++b\cdot(d_s-y^a_s)^+-h\cdot(y^*_{f,\beta}-d_s)^+-b\cdot(d_s-y^*_{f,\beta})^+]\},
\end{equation}
where $\mbox{AVG}$ stands for average over the $L=100$ demand paths.

For any $\alpha\in (0,1)$, we let $R^{a,t}_{\alpha,\beta}$ be the conditional value at risk at the $\alpha$-quantile of the $K$ regrets $r^{a,t}_{f,\beta}$. For instance, $R^{a,100}_{95\%,\beta}$ would stand for the average of the top 50 highest $r^{a,100}_{f,\beta}$ values, where each is the regret of policy $a$ under the designated $\beta$ value by time $t=100$, out of the $K=1,000$ randomly generated $f$'s. Also, $R^{a,t}_{0\%,\beta}$ would be the average of all the $r^{a,100}_{f,\beta}$'s of $f$'s sampled from all over $\Delta$. In addition, $R^{a,t}_{99.9\%,\beta}$ would be the worst average regret found over the $K=1,000$ randomly generated distributions $f$. Suppose policy $a$ is the ${\bf y}$ studied earlier. Then, when $K$ and $L$ both approach $+\infty$ and $\alpha$ approaches 100\%, the value $R^{a,t}_{\alpha,\beta}$ will approach $\sup_{f\in \Delta}R^t_f({\bf y})$, the focal point of Theorem~\ref{newsboy-up-hp}, at the particular $\beta$. Since $L$ is finite, each  $r^{a,t}_{f,\beta}$ is merely an approximation of the regret $R^t_f({\bf y})$. Moreover, the finiteness of $K$ means that the $f$ in $\Delta$ generating the worst regret will most likely be missed. Nevertheless, the $R^{a,t}_{\alpha,\beta}$ values will yield insights into regrets stemming from policies $a$.

We mainly test two policies $a$, with $a=0$ and 1, respectively. Policy 0 is our newsvendor-based one defined through~(\ref{yt-def}) and~(\ref{newsboy}). Policy 1 is adapted from the SA-based one proposed by Huh and Rusmevichientong \cite{HR09} to the current nonperishable-inventory setting. It still follows~(\ref{yt-def}), although its generation of the $\hat y_t$'s is different from~(\ref{newsboy}). For the latter, let $\varepsilon_t=\bar d/((h\vee b)\cdot\sqrt{t})$. There is also an auxiliary process $\hat z_t$. In the beginning, $y_1=\hat y_1=\hat z_1=0$. Then for $t=2,3,...$,
\begin{equation}\label{lala}
\hat y_t=\left\{\begin{array}{ll}
\lfloor \hat z_t\rfloor, & \mbox{ with probability }\lceil \hat z_t\rceil-\hat z_t,\\
\lceil \hat z_t\rceil, & \mbox{ with probability }1-(\lceil \hat z_t\rceil-\hat z_t),
\end{array}\right.\end{equation}
where
\begin{equation}\label{lalala}
\hat z_t=\left\{\begin{array}{ll}
(\hat z_{t-1}-h\varepsilon_{t-1})\vee 0\wedge \bar d, & \mbox{ when }\begin{array}{l}
\hat y_{t-1}=\lfloor \hat z_{t-1}\rfloor\mbox{ and }d_{t-1}\leq y_{t-1},\mbox{ or }\\
\lfloor \hat z_{t-1}\rfloor<\hat y_{t-1}=\lceil\hat z_{t-1}\rceil\mbox{ and }d_{t-1}\leq y_{t-1}-1,
\end{array}\\
(\hat z_{t-1}+b\varepsilon_{t-1})\vee 0\wedge \bar d, & \mbox{ when }\begin{array}{l}
\hat y_{t-1}=\lfloor \hat z_{t-1}\rfloor\mbox{ and }d_{t-1}\geq y_{t-1}+1,\mbox{ or }\\
\lfloor \hat z_{t-1}\rfloor<\hat y_{t-1}=\lceil\hat z_{t-1}\rceil\mbox{ and }d_{t-1}\geq y_{t-1}.
\end{array}\end{array}\right.\end{equation}

%Typical $\alpha$'s are 0\%, 95\%, and 99.9\%. Typical $t$ values are 100, 200, ..., 10,000.
We have considered policy 2, which uses both~(\ref{yt-def}) and the definition of the $\varepsilon_t$ sequence. The policy is inspired by both stochastic approximation and Derman's \cite{D57} up-and-down idea. Instead of~(\ref{newsboy}) or~(\ref{lala}) and~(\ref{lalala}), it uses the following for its updating of $\hat y_t$:
\begin{equation}
\hat y_t=\left\{\begin{array}{ll}
\hat y_t-1\mbox{ with probability }h\varepsilon_{t-1}, & \mbox{ when }d_{t-1}\leq y_{t-1}-1,\\
\hat y_{t-1}+1\mbox{ with probability }b\varepsilon_{t-1}, & \mbox{ when }d_{t-1}\geq y_{t-1}+1,\\
\hat y_{t-1}-\mbox{sgn}(h-b)\mbox{ with probability }|h-b|\cdot \varepsilon_{t-1}/2, & \mbox{ when }d_{t-1}=y_{t-1},\\
\hat y_{t-1}, & \mbox{ otherwise}.
\end{array}\right.\end{equation}
However, our simulation study indicates that policy 2 is not competitive against either of the previous two policies, except when $\beta$ is close to 0.5, at which time it is better than policy 1 in some occasions. So we omit presenting its performances.

\begin{figure}[ht]
\centering
\includegraphics[scale=0.7]{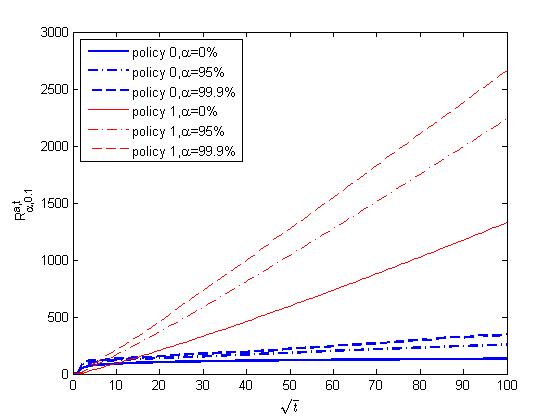}
\caption{$R^{a,t}_{\alpha,\beta}$ Values when $\beta=0.1$}\label{fig1}
\end{figure}

\begin{figure}[ht]
\centering
\includegraphics[scale=0.7]{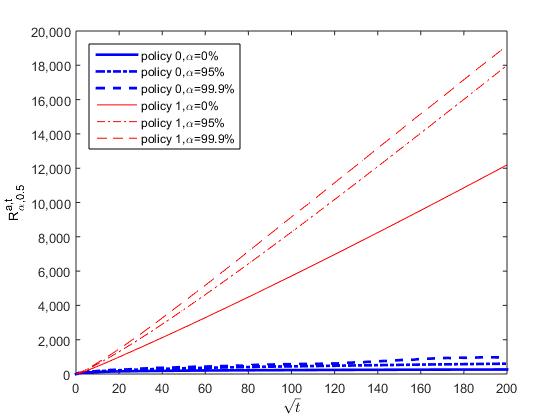}
\caption{$R^{a,t}_{\alpha,\beta}$ Values when $\beta=0.5$}\label{fig5}
\end{figure}

\begin{figure}[ht]
\centering
\includegraphics[scale=0.7]{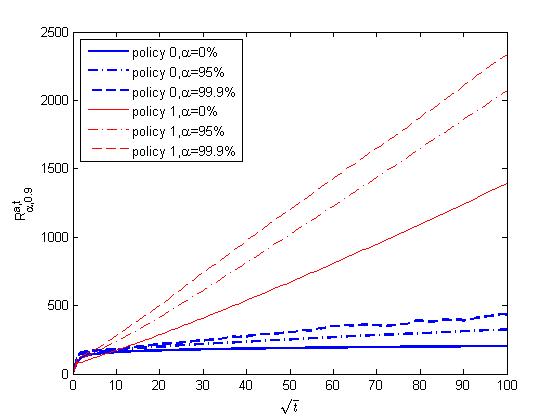}
\caption{$R^{a,t}_{\alpha,\beta}$ Values when $\beta=0.9$}\label{fig09}
\end{figure}

At various $\alpha$ and $\beta$ values, and at different $t$ points, we compare $R^{a,t}_{\alpha,\beta}$ among policies. At various $\beta$ values, we can evaluate $R^{a,t}_{\alpha,\beta}$ for policies $a=0$ and 1 at times $t=1^2$, $2^2$, $...$, $100^2$ and $\alpha$ values at 0\%, 95\%, and 99.9\%. Except for scaling differences, we find the basic findings do not depend much on $\beta$. In Figures~\ref{fig1} to~\ref{fig09}, we just present results at the three $\beta$ values of 0.1, 0.5, and 0.9. Instead of $t$, we have used $\sqrt{t}$ as our horizontal axis.

%\begin{figure}[ht]
%\centering
%\includegraphics[scale=0.7]{b03.jpg}
%\caption{$R^{a,t}_{\alpha,\beta}$ Values when %$\beta=0.3$}\label{fig3}
%\end{figure}

%\begin{figure}[ht]
%\centering
%\includegraphics[scale=0.7]{b07.jpg}
%\caption{$R^{a,t}_{\alpha,\beta}$ Values when %$\beta=0.7$}\label{fig7}
%\end{figure}

\begin{figure}[ht]
	\centering
	\includegraphics[scale=0.7]{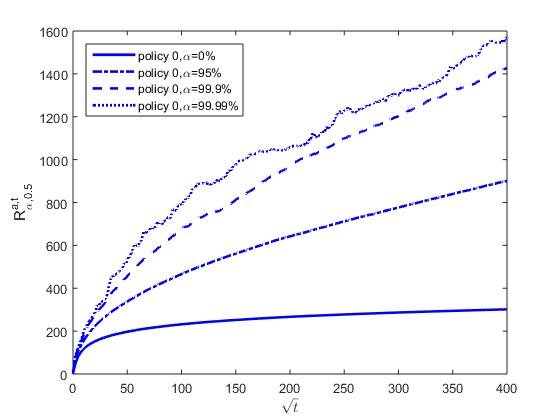}
	\caption{$R^{0,t}_{\alpha,\beta}$ Values when $\beta=0.5$}\label{fig5aa}
\end{figure}

From these figures, we can observe that policy 0 generates much smaller regrets, in both average and worst senses, than policy 1. This should be anticipated, as policy 0 utilizes more information regarding past observations than policy 1. We also see that regrets for both policies grow at approximately the rate of $\sqrt{t}$. In Figure~\ref{fig5aa}, we provide a close-up for policy 0 at $\beta=0.5$. Here, we have sampled $K=10,000$ demand distributions, made measurements at time points $t=1^2,2^2,...,400^2$, and tried $\alpha=0\%$, $95\%$, $99.9\%$, and $99.99\%$. 

%\begin{figure}[ht]
%\centering
%\includegraphics[scale=0.7]{newfig4_2.jpg}
%\caption{$R^{0,t}_{\alpha,\beta}$ Values when %$\beta=0.5$}\label{fig10}
%\end{figure}

\begin{figure}[ht]
	\centering
	\includegraphics[scale=0.7]{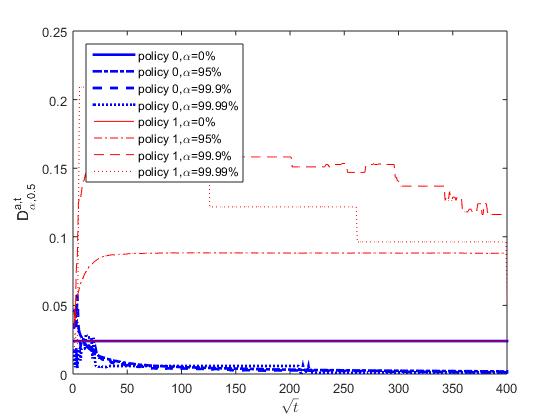}
	\caption{$D^{a,t}_{\alpha,\beta}$ Values when $\beta=0.5$}\label{fig5a}
\end{figure}

%The consistent $\sqrt{t}$-paced regret growths provide strong motivation for future research to see if the $\epsilon$-term in Theorem~\ref{newsboy-up-hp} can be eradicated for our policy 0. 
To understand what roles the distributions' separations from $\beta$ have played in the formation of regrets, let us define $D^{a,t}_{\alpha,\beta}$ as the average of the separations $\delta_{f,\beta}$, as defined by~(\ref{ag-def}) and~(\ref{small-def}), from among the $K\cdot (1-\alpha)$ distributions $f$ with the worst regrets. For instance, $D^{a,100}_{95\%,\beta}$ would be the average of the $\delta_{f,\beta}$'s among the 500 $f$'s which give the worst $r^{a,100}_{f,\beta}$ by time $t=100$, out of the $K=10,000$ randomly generated $f$'s. In Figure~\ref{fig5a}, we present $D^{a,t}_{\alpha,\beta}$ for policies $a=0$ and 1 at times $t=1^2, 2^2,...,400^2$, $\alpha$ values at 0\%, 95\%, 99.9\%, and 99.99\%, and $\beta=0.5$. Pictures at other $\beta$ values look similar. %So we omit presenting them.

At $\alpha=0\%$, the values $D^{a,t}_{0\%,0.5}$ are flat when $t$ changes, because it is always the average of the $K=10,000$ $\delta_{f,0.5}$'s. From Figure~\ref{fig5a}, we also see that $D^{0,t}_{\alpha,0.5}$ is decreasing in $\alpha$ at large enough $t$'s. This is consistent with Theorem~\ref{newsboy-up}, showing that, in the long run, the selection of distributions $f$ with large regrets will gravitate toward those with small separations from $\beta=0.5$. In contrast, $D^{1,t}_{\alpha,0.5}$ seems to receive no clear influence from $\alpha$. %y 1. These two trends combine to produce the phenomenon that $D^{0,t}_{\alpha,0.5}$ is below $D^{1,t}_{\alpha,0.5}$, with the gap widening as $\alpha$ increases. %This again attests to the emphasis that policy 0

%he best simulated regret measured at different levels of $\alpha$. The regret grows at an approximate rate of $\sqrt{t}$ under policy 1 whereas a much slower rate under policy 0. The slower growth rate under policy 0 is probably due to the constant upper bound when the demand distribution  has certain separation with $\beta$.

%Figure ~\ref{fig9} and Figure ~\ref{fig10} show that in the early stages, the regret has a dramatic growth. As time goes on, the growth in the regret becomes  more and more steady and is more or less linear with $\sqrt{t}$. It is probably due to the great variability in the choice of the ordering up to quantity as there is not much information of demand available. The deviation of the simulated regret from the expected regret may also contribute to the dramatic growth. As time goes on, the algorithem collect more and more information of demand, the variability in the ordering up to quantity becomes lower, and therefore shows perhaps the worst regret which grows almost linearly with $\sqrt{t}$.
	
\begin{figure}[ht]
	\centering
		\includegraphics[scale=0.7]{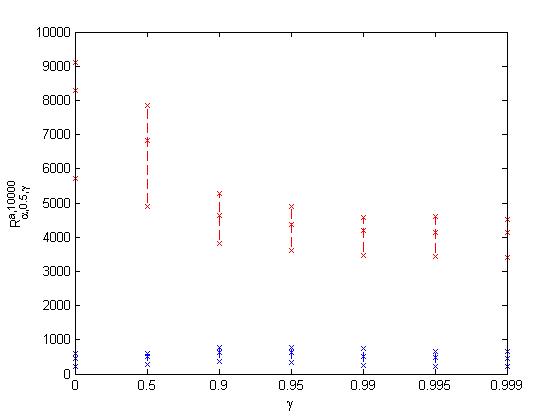}
	\caption{$R^{a,t}_{\alpha,\beta,\gamma}$ Values when $\beta=0.5$ and $t=10000$}\label{f-g}
\end{figure}

We can also introduce an inseparability index $\gamma\in [0,1)$ to indicate how difficult it is to separate demand distributions $f$ from $\beta$. Suppose $d=1,2,...,\bar d,\bar d+1$ with $\xi_{(d-1)}<\beta<\xi_{(d)}$ has been identified. Then, in the place of~(\ref{replacable}), we generate $\eta_1,...,\eta_{d-1}$ so that
\begin{equation}\label{rep1}
\eta_i=\frac{\xi_{(d-1)}+\gamma\cdot(\beta-\xi_{(d-1)})}{\xi_{(d-1)}}\cdot\xi_{(i)},\hspace*{.8in}\forall i=1,2,...,d-1,
\end{equation}
and generate $\eta_d,...,\eta_{\bar d}$ so that
\begin{equation}\label{rep2}
\eta_i=1-\frac{1-\xi_{(d)}+\gamma\cdot(\xi_{(d)}-\beta)}{1-\xi_{(d)}}\cdot (1-\xi_{(i)}),\hspace*{.8in}\forall i=d,d+1,...,\bar d.
\end{equation}
All other steps are the same as before. Our previous case happens when $\gamma$ is set at the default value 0. When $\gamma$ grows, there will be more chance that the generated $f$ has a smaller separation from $\beta$. We can define $R^{a,t}_{\alpha,\beta,\gamma}$ as the $\alpha$-quantile of the $K$ regrets $r^{a,t}_{f,\beta}$, but this time with the $f$'s generated under the given $\gamma$ using~(\ref{rep1}) and~(\ref{rep2}) instead of~(\ref{replacable}). The previous $R^{a,t}_{\alpha,\beta}$ is just $R^{a,t}_{\alpha,\beta,0}$. In Figure~\ref{f-g}, we show $R^{a,10000}_{\alpha,0.5,\gamma}$ for policies $a=0$ and 1 at  $\alpha$ values 0\%, $95\%$, and $99.9\%$, and $\gamma$ values at 0, 0.5, 0.9, 0.95, 0.99, 0.995, and 0.999.

Figure~\ref{f-g} shows that $R^{1,10000}_{\alpha,0.5,\gamma}$ is decreasing with $\gamma$. This is consistent with our earlier observation that smaller separations seem to help improve the performance of policy 1. At the same time, the increase of $\gamma$ does not make policy 0 significantly worse. This seems to suggest a slow rise of the constant bound in Theorem~\ref{newsboy-up} when the smallest allowed level of separation $\epsilon$ dwindles. For this to be reconcilable with the $\sqrt{t}$-sized rise in regret as indicated by both Theorem~\ref{newsboy-up-hp} and Figure~\ref{fig5aa}, we will need the ``leading'' separation for policy 0 to decrease over time. This has somewhat been confirmed by Figure~\ref{fig5a}. %Overall, policy 0 will remain competitive even when separations become extremely small.

\section{Concluding Remarks}\label{conclusions}

In regret bounds for a newsvendor-based adaptive policy, we have contributed to inventory control involving unknown demand of discrete nonperishable items. Currently, our analysis relies on knowledge of the maximum per-period demand $\bar d$. Also, our universal bound is related to a strictly positive parameter $\epsilon$. Both deserve more attention. In addition, the newsvendor-based policy requires higher observability of historical demand levels than other policies, say the SA-based one. This makes it ill suited to situations involving demand censoring. Furthermore, we have not touched on realistic features like nonzero setup costs or lead times. 

 So a great deal awaits to be done in future research.

% Other policies are certainly worth investigating. But the more realistic discrete-demand setting might pose as a hurdle. We know stochastic approximation, due to its fine granularity, works well for the continuous setting. But its counterpart in the discrete setting, the up-and-down method, does not lead to convergence. Instead, one has to resort to the most frequently encountered solution.

%\vspace*{.8in}
\newpage
%\def\setpagenumber#1{\global\setcounter{page}{#1}}
%\setcounter{secnumdepth}{1}
%\setpagenumber{1}

\noindent{\bf\Large Appendices}%\vspace*{.05in}
\appendix
\numberwithin{equation}{section}

\section{Proofs of Section~\ref{1-bound}}\label{appendix1}

\noindent{\bf Proof of Proposition~\ref{1-bound}: }Combining~(\ref{def0}) and~(\ref{def1}), we have
\begin{equation}\label{okkk}\begin{array}{l}
R^{T1}_f({\bf y})=h\cdot  \sum_{t=1}^{T}\{ \mathbb{E}_f[(\hat y_t-D_t)^+]-\mathbb{E}_f[(y^*_{f,\beta}-D_t)^+]\}\\
\;\;\;\;\;\;\;\;\;\;\;\;\;\;\;\;\;\;+ b\cdot\sum_{t=1}^{T} \{\mathbb{E}_f[(D_t-\hat y_t)^+]-\mathbb{E}_f[(D_t-y^*_{f,\beta})^+]\}.
\end{array}\end{equation}
When $x\geq y$,
\begin{equation}\label{ineq10}
(x-z)^+-(y-z)^+=(x-\max\{y,z\})^+;
\end{equation}
whereas, when $x\leq y$,
\begin{equation}\label{ineq20}
(z-x)^+-(z-y)^+=(\min\{y,z\}-x)^+.
\end{equation}
Plugging these into~(\ref{okkk}), we obtain
\begin{equation}\begin{array}{l}
R^{T1}_f({\bf y})=h\cdot\sum_{t=1}^T \sum_{y=y^*_{f,\beta}+1}^{\bar d} \mathbb{P}_f[\hat y_t=y]\cdot\mathbb{E}_f[(y-\max\{y^*_{f,\beta},D_t\})^+]\\
\;\;\;\;\;\;\;\;\;\;\;\;\;\;\;\;\;\;+b\cdot\sum_{t=1}^T \sum_{y=0}^{y^*_{f,\beta}-1} \mathbb{P}_f[\hat y_t=y]\cdot\mathbb{E}_f[(\min\{y^*_{f,\beta},D_t\}-y)^+],
\end{array}\end{equation}
which is below $(h\bar d+b\bar d)\cdot \sum_{t=1}^{T}\mathbb{P}_f[\hat y_t\geq y^*_{f,\beta}+1]\vee \mathbb{P}_f[\hat y_t\leq y^*_{f,\beta}-1]$. Therefore,
\begin{equation} R^{T1}_f({\bf y})\leq (h\bar d+b\bar d)\cdot (\sum_{t=1}^{\tau_{f,\beta}}+\sum_{t=\tau_{f,\beta}+1}^{+\infty})\mathbb{P}_f[\hat y_t\geq y^*_{f,\beta}+1]\vee \mathbb{P}_f[\hat y_t\leq y^*_{f,\beta}-1],\end{equation}
which, due to~(\ref{useful1}) to~(\ref{aha1}), is below the desired upper bound. \qed %$(h\bar d+b\bar d)\cdot (\tau_{f,\beta}+1/(1-\exp(-\kappa_{f,\beta}/2)))/2$. \qed

\noindent{\bf Proof of Proposition~\ref{2-bound}: }Combining~(\ref{def0}) and~(\ref{def2}), we have
\begin{equation}R^{T2}_f({\bf y})=h\cdot \sum_{t=3}^{T}\mathbb{E}_f[(y_t-D_t)^+-(\hat y_t-D_t)^+]
+b \cdot\sum_{t=3}^{T}\mathbb{E}_f[(D_t-y_t)^+-(D_t-\hat y_t)^+].\end{equation}
Note $y_t\geq \hat y_t$ from~(\ref{yt-def}). So in view of both~(\ref{ineq10}) and the fact that $(d-y_t)^+-(d-\hat y_t)^+\leq 0$,
\begin{equation}R^{T2}_f({\bf y})\leq h\cdot  \sum_{t=3}^{T}\mathbb{E}_f[(y_t-\max\{ \hat y_t,D_t\})^+].\end{equation}
But this is further below $h\cdot \sum_{t=3}^{T}\mathbb{E}_f[(y_t-\hat y_t)^+]$. Hence,
\begin{equation}\label{kooo}
R^{T2}_f({\bf y})\leq h\bar d\cdot\tau_{f,\beta}+h\bar d\cdot\sum_{t=\tau_{f,\beta}}^{+\infty}\mathbb{P}_f[y_{t+1}\geq\hat y_{t+1}+1]. \end{equation}
So the key is to bound $\mathbb{P}_f[y_{t+1}\geq\hat y_{t+1}+1]$ for $t\geq\tau_{f,\beta}$. However,
\begin{equation}\begin{array}{l}
\mathbb{P}_f[y_{t+1}\geq\hat y_{t+1}+1] =\mathbb{P}_f[y_{t+1}\geq \hat{y}_{t+1}+1|\hat{y}_{t+1}=y^*_{f,\beta}]\cdot\mathbb{P}_f[\hat{y}_{t+1}=y^*_{f,\beta}]\\
\;\;\;\;\;\;\;\;\;\;\;\;\;\;\;\;\;\;+\mathbb{P}_f[y_{t+1}\geq \hat{y}_{t+1}+1 |\hat{y}_{t+1} \neq y^*_{f,\beta}]\cdot\mathbb{P}_f[\hat{y}_{t+1} \neq y^*_{f,\beta}]\\
\;\;\;\;\;\;\leq \mathbb{P}_f[y_{t+1}\geq y^*_{f,\beta}+1]+\mathbb{P}_f[\hat y_{t+1}\leq y^*_{f,\beta}-1]+\mathbb{P}_f[\hat y_{t+1}\geq y^*_{f,\beta}+1].
\end{array}\end{equation}
Thus, from~(\ref{useful1}) to~(\ref{aha1}), 
\begin{equation}\label{18}
\mathbb{P}_f[y_{t+1}\geq\hat y_{t+1}+1] \leq \mathbb{P}_f[y_{t+1}\geq y^*_{f,\beta}+1]+\exp(-\frac{\kappa_{f,\beta}\cdot(t-\tau_{f,\beta})}{2}).
\end{equation}
Meanwhile,
\begin{equation}\label{19}\begin{array}{l}
\mathbb{P}_f[y_{t+1}\geq y^*_{f,\beta}+1] = \mathbb{P}_f[y_{t+1}\geq y^*_{f,\beta}+1|y_{t}\geq y^*_{f,\beta}+1]\cdot \mathbb{P}_f[y_{t}\geq y^*_{f,\beta}+1]\\
\;\;\;\;\;\;\;\;\;\;\;\;\;\;\;+\mathbb{P}_f[y_{t+1}\geq y^*_{f,\beta}+1|y_{t}\leq y^*_{f,\beta}]\cdot\mathbb{P}_f[y_{t}\leq y^*_{f,\beta}]\\
\;\;\;\;\;\;\leq\mathbb{P}_f[y_{t}-D_{t}\geq y^*_{f,\beta}+1|y_{t}\geq y^*_{f,\beta}+1]\cdot \mathbb{P}_f[y_{t}\geq y^*_{f,\beta}+1]\\
\;\;\;\;\;\;\;\;\;\;\;\;\;\;\;+\mathbb{P}_f[\hat{y}_{t+1}\geq y^*_{f,\beta}+1| y_{t}\geq y^*_{f,\beta}+1]\cdot \mathbb{P}_f[y_{t}\geq y^*_{f,\beta}+1]\\
\;\;\;\;\;\;\;\;\;\;\;\;\;\;\;+\mathbb{P}_f[\hat{y}_{t+1}\geq y^*_{f,\beta}+1| y_{t}\leq y^*_{f,\beta}]\cdot \mathbb{P}_f[y_{t}\leq y^*_{f,\beta}]\\
\;\;\;\;\;\;=\mathbb{P}_f[y_{t}-D_{t}\geq y^*_{f,\beta}+1|y_{t}\geq y^*_{f,\beta}+1]\cdot \mathbb{P}_f[y_{t}\geq y^*_{f,\beta}+1]+\mathbb{P}_f[\hat{y}_{t+1}\geq y^*_{f,\beta}+1],
\end{array}\end{equation}
where the first equality is an identity, the first inequality comes from~(\ref{yt-def}), and the next equality is another identity when combining the previous two terms. Note $y_{t+1}=y_t-D_t$ or $y_{t+1}=\hat y_{t+1}$, and the latter would have occurred for sure when $y_{t+1}\geq y_t+1$. Concerning part of the first term in the above, as $y_t\leq \bar d$,
\begin{equation}\label{191}
\mathbb{P}_f[y_{t}-D_{t}\geq y^*_{f,\beta}+1|y_{t}\geq y^*_{f,\beta}+1]\leq \mathbb{P}_f[D_t\leq \bar d-y^*_{f,\beta}-1]=F_f(\bar d-y^*_{f,\beta}-1).
\end{equation}
Also in view of~(\ref{useful1}) to~(\ref{aha1}), we have from~(\ref{19}) and~(\ref{191}) that
\begin{equation} \mathbb{P}_f[y_{t+1}\geq y^*_{f,\beta}+1]\leq F_f(\bar d-y^*_{f,\beta}-1)\cdot\mathbb{P}_f[y_{t}\geq y^*_{f,\beta}+1]+\frac{\exp(-\kappa_{f,\beta}\cdot(t-\tau_{f,\beta})/2)}{2}.\end{equation}
Because $\epsilon_f\equiv f(\bar d)>0$,
\begin{equation} F_f(\bar d-y^*_{f,\beta}-1)\leq F_f(\bar d-1)=1-f(\bar d)= 1-\epsilon_f<1.\end{equation}
So for $t\geq \tau_{f,\beta}$, we have the iterative relation
\begin{equation} \mathbb{P}_f[y_{t+1}\geq y^*_{f,\beta}+1]\leq (1-\epsilon_f)\cdot \mathbb{P}_f[y_t\geq y^*_{f,\beta}+1]+\frac{\exp(-\kappa_{f,\beta}\cdot(t-\tau_{f,\beta})/2)}{2}.\end{equation}
Noting that $\mathbb{P}_f[y_{\tau_{f,\beta}}\geq y^*_{f,\beta}+1]\leq 1$, we obtain
\begin{equation} \mathbb{P}_f[y_{t+1}\geq y^*_{f,\beta}+1]\leq (1-\epsilon_f)^{t-\tau_{f,\beta}+1}+\frac{1}{2}\cdot\sum_{s=0}^{t-\tau_{f,\beta}}(1-\epsilon_f)^{t-\tau_{f,\beta}-s}\cdot\exp(-\frac{\kappa_{f,\beta}\cdot s}{2}).\end{equation}
Summing over $t=\tau_{f,\beta},\tau_{f,\beta}+1,...$, we obtain
\begin{equation}\label{20}
\sum_{t=\tau_{f,\beta}}^{+\infty} \mathbb{P}_f[y_{t+1}\geq y^*_{f,\beta}+1]
\leq  \frac{1-\epsilon_f}{\epsilon_f}+\frac{1}{2\epsilon_f\cdot(1-\exp(-\kappa_{f,\beta}/2))}.
\end{equation}

Combining~(\ref{kooo}),~(\ref{18}), and~(\ref{20}), we obtain
\begin{equation}\begin{array}{ll}
R^{T2}_f({\bf y})&\leq h\bar d\cdot\tau_{f,\beta}+h\bar d\cdot\sum_{t=\tau_{f,\beta}}^{+\infty}\mathbb{P}_f[y_{t+1}\geq y^*_{f,\beta}+1]\\
&\;\;\;\;\;\;+h\bar d\cdot\sum_{t=\tau_{f,\beta}}^{+\infty}\exp(-\kappa_{f,\beta}\cdot(t-\tau_{f,\beta})/2)\\
&\leq h\bar d\cdot\tau_{f,\beta}+h\bar d\cdot\sum_{t=\tau_{f,\beta}}^{+\infty}\mathbb{P}_f[y_{t+1}\geq y^*_{f,\beta}+1]+ h\bar d/(1-\exp(-\kappa_{f,\beta}/2)),
\end{array}\end{equation}
which is below the desired upper bound. \qed %$h\bar

\noindent{\bf Proof of Theorem~\ref{newsboy-up}: }Putting~(\ref{regret0}) as well as Propositions~\ref{1-bound} and~\ref{2-bound} together, we can obtain an upper bound for $R^T_f({\bf y})$:
\begin{equation}\label{qqq}
(2h\bar d+b\bar d)\cdot\tau_{f,\beta}+\frac{3h\bar{d}+b\bar{d}}{2}\cdot\frac{1}{1-\exp(-\kappa_{f,\beta}/2)}
+h\bar d\cdot(\frac{1-\epsilon_f}{\epsilon_f}+\frac{1}{2\epsilon_f\cdot(1-\exp(-\kappa_{f,\beta}/2))}).
\end{equation}
%When case 2 occurs, let $y^1_{f,\beta}\geq y^*_{f,\beta}$ be such that
%\begin{equation}\label{posi} \beta=F_f(y^*_{f,\beta})=F_f(y^*_{f,\beta}+1)=\cdots=F_f(y^1_{f,\beta})<F_f(y^1_{f,\beta}+1).\end{equation}
%From~(\ref{cond1}), we see that
%\begin{equation} Q_f(y^*_{f,\beta})=Q_f(y^*_{f,\beta}+1)=\cdots=Q_f(y^1_{f,\beta}+1)<Q_f(y^1_{f,\beta}+2).\end{equation}
%So instead of estimating $\mathbb{P}_f[\hat y_t\geq y^*_{f,\beta}+1]$, we need only to estimate $\mathbb{P}_f[\hat y_t\geq y^1_{f,\beta}+2]$. But according to~(\ref{ineq1}), this is the same as $\mathbb{P}_{\gamma_{f,\beta}}[\hat\gamma_{t-1}<\beta]$ for $\gamma_{f,\beta}=F_f(y^1_{f,\beta}+1)$. Now due to~(\ref{posi}),  $\min\{\beta-\alpha_{f,\beta},\gamma_{f,\beta}-\beta\}=\min\{f(y^*_{f,\beta}),f(y^1_{f,\beta}+1)\}>0$. So the same bounds as in Propositions~\ref{1-bound} and~\ref{2-bound} can be established.
For $f\in\Delta_{\epsilon,\delta,\beta}$, we already have $\epsilon_f\geq \epsilon$ by the definition in~(\ref{delta-def}).
%On the other hand, a definition suitable for both cases is that
%\begin{equation}
%\alpha_{f,\beta}=\max\{F_f(d)|F_f(d)<\beta\},\hspace*{.5in}\gamma_{f,\beta}=\min\{F_f(d)|F_f(d)>\beta\}.
%\end{equation}
Also, note that $\mbox{D}_{KL}(u||\cdot)$ is convex with minimum achieved at $u$ (Cover and Thomas \cite{CT06}, Theorems 2.6.3 and 2.7.2). So in view of~(\ref{kappa-def}),~(\ref{ag-def}), and~(\ref{small-def}),
\begin{equation}
\kappa_{f,\beta}\equiv \min\{\mbox{D}_{KL}(\beta||\alpha_{f.\beta}),\mbox{D}_{KL}(\beta||\gamma_{f,\beta})\}\geq \min\{\mbox{D}_{KL}(\beta||\beta-\delta_{f,\beta}),\mbox{D}_{KL}(\beta||\beta+\delta_{f,\beta})\},
\end{equation}
which is further greater than $\kappa'_{\delta,\beta}\equiv \min\{\mbox{D}_{KL}(\beta||\beta-\delta),\mbox{D}_{KL}(\beta||\beta+\delta)\}$ due to $f$'s membership in the $\Delta_{\epsilon,\delta,\beta}$ defined by~(\ref{delta-def}). We can define $\tau'_{\delta,\beta}$ for $\kappa'_{\delta,\beta}$ in the same fashion in which $\tau_{f,\beta}$ is defined for $\kappa_{f,\beta}$; namely, through~(\ref{aha2}) and~(\ref{aha1}). Clearly, $\tau'_{\delta,\beta}\geq \tau_{f,\beta}$. Since~(\ref{qqq}) is decreasing in $\epsilon_f$, $\kappa_{f,\beta}$, and increasing in $\tau_{f,\beta}$, we can replace them by, respectively, $\epsilon$, $\kappa'_{\delta,\beta}$, and $\tau'_{\delta,\beta}$. The new right-hand side would constitute the desired  $A_{\epsilon,\delta,\beta}$.  \qed

\section{Proofs of Section~\ref{2-bound}}\label{appendix2}

\noindent{\bf Proof of Proposition~\ref{1-bound-hp}: }For $f,g\in \Delta$, note that $Q_f(y^*_{g,\beta})-Q_f(y^*_{f,\beta})$ can be written as
\begin{equation}\label{discuss} [Q_f(y^*_{g,\beta})-Q_g(y^*_{g,\beta})]+[Q_g(y^*_{g,\beta})-Q_g(y^*_{f,\beta})]+[Q_g(y^*_{f,\beta})-Q_f(y^*_{f,\beta})].
\end{equation}
While the first and third terms can be made small when $f$ and $g$ are close, the second term is always negative due to $y^*_{g,\beta}$'s optimality when the underlying demand distribution is $g$. Let us investigate how small the first and third terms can be. For any $y=0,1,...,\bar d$,
\begin{equation}\begin{array}{ll}
Q_f(y)-Q_g(y) &=h\cdot\sum_{d=0}^{y-1} [f(d)-g(d)](y-d)+b\cdot\sum_{d=y+1}^{\bar d} [f(d)-g(d)](d-y)\\
&\leq (h\vee b)\cdot\bar d\cdot \mid\mid f-g\mid\mid_1=2\cdot (h\vee b)\cdot\bar d\cdot\delta_V(f,g).
\end{array}\end{equation}
In view of the discussion around~(\ref{discuss}),
\begin{equation}\label{pre-p}
Q_f(y^*_{g,\beta})-Q_f(y^*_{f,\beta})\leq 4\cdot (h\vee b)\cdot \bar d\cdot \delta_V(f,g).
\end{equation}
Through Pinsker's inequality~(\ref{pinsker}), we see that~(\ref{pre-p}) will become
\begin{equation}\label{pinsker-son}
Q_f(y^*_{g,\beta})-Q_f(y^*_{f,\beta})\leq 2\cdot(h\vee b)\cdot\bar d\cdot \sqrt{2\mbox{D}_{KL}(g||f)}.
\end{equation}
Note~(\ref{def0}) also leads to
\begin{equation}\label{rough-son}
Q_f(y^*_{g,\beta})-Q_f(y^*_{f,\beta})\leq (h\vee b)\cdot \bar d.
\end{equation}

%Show that $Q_f(y^*_g)$ is continuous in $g$ when $g$ is around $f$ under certain metric for distributions, such as variational or Kullback-Leibler.

%Show that $Q^*_f$ is continuous in $f$ under the same metric for distributions.

For a fixed $\varepsilon>0$, consider $G=\{g\in \Delta| \mbox{D}_{KL}(g||f)\geq \varepsilon\}$. Then~(\ref{ssnv}) will result with
\begin{equation}\label{sanov-son}
\mathbb{P}_f[\mbox{D}_{KL}(\hat f_{t-1}||f)\geq \varepsilon]\leq t^{\bar d+1}\cdot\exp(-\varepsilon\cdot(t-1)).
\end{equation}
Consider $R^{T1}_f({\bf y})$ defined at~(\ref{def1}). By~(\ref{newsboy}), we have
\begin{equation} R^{T1}_f({\bf y})=\sum_{t=1}^T \{\mathbb{E}_f[Q_f(y^*_{\hat f_{t-1},\beta})]-Q_f(y^*_{f,\beta})\}.\end{equation}
Let $\varepsilon_t$ be a sequence of positive constants. We then see that $R^{T1}_f({\bf y})$ is below
\begin{equation}\label{abv}\begin{array}{l}
\sum_{t=1}^T \{\mathbb{P}_f[\mbox{D}_{KL}(\hat f_{t-1}||f)<\varepsilon_t]\cdot \mathbb{E}_f[Q_f(y^*_{\hat f_{t-1},\beta})-Q_f(y^*_{f,\beta})|\mbox{D}_{KL}(\hat f_{t-1}||f)<\varepsilon_t]\\
\;\;\;\;\;\;\;\;\;\;\;\;+\mathbb{P}_f[\mbox{D}_{KL}(\hat f_{t-1}||f)\geq\varepsilon_t]\cdot \mathbb{E}_f[Q_f(y^*_{\hat f_{t-1},\beta})-Q_f(y^*_{f,\beta})|\mbox{D}_{KL}(\hat f_{t-1}||f)\geq\varepsilon_t]\}\\
\;\;\;\;\;\;\leq (h\vee b)\cdot \bar d\cdot\sum_{t=1}^T [2\sqrt{2\varepsilon_t}+t^{\bar d+1}\cdot\exp(-\varepsilon_t\cdot (t-1))],
\end{array}\end{equation}
where the inequality comes from~(\ref{pinsker-son}) to~(\ref{sanov-son}).

%We attempt to find as good a sequence $\epsilon_t$ as possible to achieve as tight an upper bound for $R^{T1}_f({\bf y})$ as possible. For the time being,
Suppose $\varepsilon_t=(t-1)^{-1+2\epsilon}$ for some $\epsilon\in (0,1/2)$. Plugging this into~(\ref{abv}), we get
\begin{equation}\label{ko1}\begin{array}{ll}
R^{T1}_f({\bf y})&\leq (h\vee b)\cdot \bar d\cdot\{2\sqrt{2}\cdot\sum_{t=0}^{T-1}t^{-1/2+\epsilon}+\sum_{t=0}^{T-1}(t+1)^{\bar d+1}\cdot \exp(-t^{2\epsilon})\}\\
&=(h\vee b)\cdot\bar d\cdot \{2\sqrt{2}+2\sqrt{2}\cdot T_1+T_2\},
\end{array}\end{equation}
where
\begin{equation}
T_1=\sum_{t=1}^{T-1}t^{-1/2+\epsilon},\;\;\;\;\;\;\mbox{ and }\;\;\;\;\;\;T_2=\sum_{t=1}^{T-1}(t+1)^{\bar d+1}\cdot\exp(-t^{2\epsilon}).
\end{equation}
Since $t^{-1/2+\epsilon}$ is decreasing in $t$,
\begin{equation}\label{ko2}
T_1\leq\int_0^{T-1}t^{-1/2+\epsilon}\cdot dt=\frac{2}{1+2\epsilon}\cdot(T-1)^{1/2+\epsilon}.
\end{equation}
Note that $(t+1)^{\bar d+1}\cdot\exp(-t^{2\epsilon})$ is positive, increasing first, and decreasing next. So
\begin{equation}
T_2\leq \int_0^{T-1}(t+1)^{\bar d+1}\cdot\exp(-t^{2\epsilon})\cdot dt+\int_1^T(t+1)^{\bar d+1}\cdot\exp(-t^{2\epsilon})\cdot dt.
\end{equation}
Since $(t+1)^{2\epsilon}-t^{2\epsilon}<1$, we have $T_2$ being below
\begin{equation}\label{ok1}\begin{array}{l}
e\cdot[\int_0^{T-1}(t+1)^{\bar d+1}\cdot\exp(-(t+1)^{2\epsilon})\cdot dt+\int_1^T(t+1)^{\bar d+1}\cdot\exp(-(t+1)^{2\epsilon})\cdot dt]\\
\;\;\;\;\;\;\leq 2e\cdot \int_1^{+\infty} t^{\bar d+1}\cdot \exp(-t^{2\epsilon})\cdot dt,
\end{array}\end{equation}
where $e$ is the natural logarithmic base. Letting $u=t^{2\epsilon}$ and hence $t=u^{1/(2\epsilon)}$, we obtain
\begin{equation}\label{ok2}
\int_1^{+\infty}t^{\bar d+1}\cdot\exp(-t^{2\epsilon})\cdot dt =\frac{1}{2\epsilon}\cdot\int_1^{+\infty}u^{(\bar d+2-2\epsilon)/(2\epsilon)}\cdot\exp(-u)\cdot du.
\end{equation}
Meanwhile, for $\alpha>0$, integral by parts has $\int_1^{+\infty}u^\alpha\cdot \exp(-u)\cdot du$ equal to
\begin{equation}\label{ok3}
e^{-1}+\alpha e^{-1}+\cdots+(\lfloor \alpha\rfloor-1)!\cdot e^{-1}+(\lfloor \alpha\rfloor)!\cdot\int_1^{+\infty}u^{\alpha-\lfloor \alpha\rfloor-1}\cdot\exp(-u)\cdot du,
\end{equation}
which is below an $\alpha$-dependent constant. Combining~(\ref{ok1}) to~(\ref{ok3}), we see that $T_2$ is below an $\epsilon$-dependent constant. Together with~(\ref{ko1}) and~(\ref{ko2}), we see that for any $\epsilon\in (0,1/2)$, there are constants $A_\epsilon$ and $B_\epsilon$ for the intended bound to be true. Furthermore, by adopting $A_\epsilon=A_{1/3}$ and $B_\epsilon=B_{1/3}$ for $\epsilon\geq 1/2$, the inequality can be maintained for every $\epsilon>0$. \qed

\noindent{\bf Proof of Proposition~\ref{2-bound-hp}: }Let $\gamma=1-f(0)$. We divide the proof into two cases, with respectively, $\gamma\in [(1-\beta)/2,1]$ and $\gamma\in [0,(1-\beta)/2)$.

Consider the first case with $\gamma\in [(1-\beta)/2,1]$. For any positive integer $\tau_T$, we show how $\mathbb{P}_f[S_{i+1}-S_i-1\geq \tau_T+1]$ can be bounded. By the definition of the $S_i$'s around~(\ref{notso}),
\begin{equation}
\hat y_{S_{i+1}-1}\leq \hat y_{S_i}-D_{S_i}-D_{S_i+1}-\cdots-D_{S_{i+1}-2}-1.
\end{equation}
Since both $\hat y_{S_i}$ and $\hat y_{S_{i+1}-1}$ are between 0 and $\bar d$, the above necessitates that
\begin{equation}\label{musashi}
D_{S_i}+D_{S_i+1}+\cdots+D_{S_{i+1}-2}\leq \bar d-1.
\end{equation}
This is only possible when there are at least $S_{i+1}-S_i-\bar d$ zeros among the $S_{i+1}-S_i-1$ demand levels $D_{S_i},D_{S_i+1},...,D_{S_{i+1}-2}$. When $S_{i+1}-S_i-1=\tau+1\geq \bar d-1$,  the latter event's chance under $f$ with $f(0)=1-\gamma$ is, by the binomial formula,
\begin{equation}\label{bazaj}
\sum_{k=\tau-\bar d+2}^{\tau+1} \frac{(\tau+1)!}{k!\cdot(\tau+1-k)!}\cdot (1-\gamma)^k\cdot \gamma^{\tau+1-k}<(\tau+1)^{\bar d}\cdot (1-\gamma)^{\tau-\bar d+2}\cdot (1+\gamma+\cdots+\gamma^{\bar d}),
\end{equation}
which is less than $(\tau+1)^{\bar d}\cdot (1-\gamma)^{\tau-\bar d+1}$. There exists $\theta_\gamma=\bar d-1,\bar d,...$ so that when $\tau\geq \theta_{\gamma}$, the aforementioned term %$(\tau+1)^{\bar  d}\cdot (1-\gamma)^{\tau-\bar d+1}$
will decrease with $\tau$. For $\tau_T\geq \theta_\gamma$, we can thus deduce that
\begin{equation}\label{import-bound}
\mathbb{P}_f[S_{i+1}-S_i-1\geq \tau_T+1]<(\tau_T+1)^{\bar d}\cdot (1-\gamma)^{\tau_T-\bar d+1}.
\end{equation}

Each of the terms in~(\ref{ee74}) is between 0 and $(h+b)\cdot \bar d$.
%Similarly, for any $t$, let $R(t)$ be the smallest $S_i-1\geq t$. A key feature is that, if $L(t)\leq t-1$,
%\begin{equation}
%L(t)=L(t-1)=\cdots=L(L(t)-1)=L(L(t)).
%\end{equation}
%By~(\ref{def2}), as well as~(\ref{great}) to~(\ref{notso}),
%\begin{equation}
%R^{T2}_f({\bf y})=\sum_{t=3}^T\mathbb{E}_f[Q_f(\hat y_{L(t)}-D_{L(t)}-\cdots-D_{t-1})-Q_f(\hat y_t)].
%\end{equation}
So by~(\ref{critical}) and~(\ref{import-bound}),
\begin{equation}\label{okkk0}\begin{array}{l}
R^{T2}_f({\bf y})\leq \sum_{t=3}^T\sum_{s=2\vee (t-\tau_T)}^{t-1}\mathbb{E}_f[\mid Q_f(\hat y_s-D_s-\cdots-D_{t-1})-Q_f(\hat y_t)\mid\times\\
\;\;\;\;\;\;\;\;\;\;\;\;\;\;\;\;\;\;\;\;\;\;\;\;\;\;\;\;\;\;\times {\bf 1}(\hat y_t
%\vee\cdots\vee(\hat y_{s+1}-d_{s+1}-\cdots-D_{t-1})
\leq \hat y_s-D_s-\cdots-D_{t-1}-1)]\\
\;\;\;\;\;\;\;\;\;\;\;\;\;\;\;\;\;\;\;\;\;\;\;\;+(T-2)\cdot (h\bar d+b\bar d)\cdot (\tau_T+1)^{\bar d}\cdot (1-\gamma)^{\tau_T-\bar d+1}.
\end{array}\end{equation}
The above right-hand side can be written as
\begin{equation}\label{kkko}
\sum_{\tau=1}^{\tau_T}R^{T2,\tau}_f({\bf y})+(T-2)\cdot (h\bar d+b\bar d)\cdot (\tau_T+1)^{\;\bar d}\cdot (1-\gamma)^{\tau_T-\bar d+1},
\end{equation}
where for $\tau=1,2,...,\tau_T$,
\begin{equation}\label{joke} \begin{array}{l}
R^{T2,\tau}_f({\bf y})=\sum_{t=\tau+2}^{T}\mathbb{E}_f[\mid Q_f(\hat y_{t-\tau}-D_{t-\tau}-\cdots-D_{t-1})-Q_f(\hat y_t)\mid\times\\
\;\;\;\;\;\;\;\;\;\;\;\;\;\;\;\;\;\;\;\;\;\;\;\;\times {\bf 1}(\hat y_t
%\vee\cdots\vee (\hat y_{t-\tau+1}-D_{t-\tau+1}-\cdots-D_{t-1})
\leq \hat y_{t-\tau}-D_{t-\tau}-\cdots-D_{t-1}-1)].
\end{array}\end{equation}
%where $s\geq t-\bar d$ stems from the fact that $\hat y_t$ and $\hat y_s$ both need to stay between 0 and $\bar d$.
%Because both $\hat y_t$ and $\hat y_{t-\tau}$ are between 0 and $\bar d$,
%\begin{equation}\begin{array}{l}
%\mathbb{P}_f[\hat y_t\vee\cdots\vee (\hat y_{t-\tau+1}-D_{t-\tau+1}-\cdots-D_{t-1})\leq \hat y_{t-\tau}-D_{t-\tau}-\cdots-D_{t-1}-1]\\
%\;\;\;\;\;\;\leq \mathbb{P}_f[D_{t-\tau}+\cdots+D_{t-1}\leq \bar d-1].
%\end{array}\end{equation}
%Since the maximum that $D_{t-\tau}+\cdots+D_{t-1}$ can reach is $\tau\bar d$, we must have
%\begin{equation} p\cdot(\bar d-1)+(1-p)\cdot\tau\bar d\geq \tau\cdot \mathbb{E}_f[D]\equiv \tau\cdot\sum_{d=0}^{\bar d}d\cdot f(d).\end{equation}
%For $\tau\geq \bar d$,
%\begin{equation} \mathbb{P}_f[D_{t-\tau}+\cdots+D_{t-1}\leq \bar d-1]\leq \mathbb{P}_f[\mbox{at least }\tau-\bar d+1\mbox{ of the demand levels are }0]\equiv p.\end{equation}
%Suppose for some $\gamma\in (0,1]$, it is know that $f(0)\leq 1-\gamma$. Then, from the above, we can conclude that
%\begin{equation}
%\sum_{t=\tau+1}^{+\infty}\mathbb{P}_f[\hat y_t\leq \hat y_{t-\tau}-D_{t-\tau}-\cdots-D_{t-1}-1]\leq \sum_{t=\tau+1}^{+\infty}\bar d\cdot \tau^{\bar d}\cdot (1-\gamma)^{\tau-\bar d+1}
%\end{equation}
%Let us bound the $R^{T2,\tau}_f({\bf y})$ terms.
By~(\ref{yf-def}) and~(\ref{newsboy}), we have $\hat y_t\leq\hat y_{t-\tau}-D_{t-\tau}-\cdots-D_{t-1}-1$ only if
\begin{equation} \hat F_{t-\tau-1}(\hat y_{t-\tau}-1)<\beta\leq \hat F_{t-1}(\hat y_t)\leq \hat F_{t-1}(\hat y_{t-\tau}-D_{t-\tau}-\cdots-D_{t-1}-1).\end{equation}
Also, due to the nature of the empirical distribution as illustrated in~(\ref{emp}),
\begin{equation} \hat F_{t-1}(\hat y_{t-\tau}-D_{t-\tau}-\cdots-D_{t-1}-1)\leq \hat F_{t-1}(\hat y_{t-\tau}-1)\leq\hat F_{t-\tau-1}(\hat y_{t-\tau}-1)+\frac{\tau}{t-\tau}.\end{equation}
Therefore, $\hat y_t\leq\hat y_{t-\tau}-D_{t-\tau}-\cdots-D_{t-1}-1$ only if
\begin{equation}\label{dada}
\beta \leq\hat F_{t-1}(\hat y_t)
\leq \hat F_{t-1}(\hat y_{t-\tau}-D_{t-\tau}-\cdots-D_{t-1}-1)<\beta+\frac{\tau}{t-\tau}\leq \beta+\frac{\tau_T}{t-\tau},
\end{equation}
an inequality alluded to earlier in~(\ref{dada-o}). On the other hand,~(\ref{def0}) has that, for $y\leq z$,
\begin{equation}\label{77o}\begin{array}{l}
Q_f(z)-Q_f(y)\\
\;\;\;=h\cdot\sum_{d=0}^{z-1}F_f(d)+b\cdot\sum_{d=z}^{\bar d-1}(1-F_f(d))-h\cdot\sum_{d=0}^{y-1}F_f(d)-b\cdot\sum_{d=y}^{\bar d-1}(1-F_f(d))\\
\;\;\;=h\cdot\sum_{d=y}^{z-1}F_f(d)-b\cdot\sum_{d=y}^{z-1}(1-F_f(d))= (h+b)\cdot\sum_{d=y}^{z-1}(F_f(d)-\beta) .
\end{array}\end{equation}
Now by~(\ref{joke}),~(\ref{dada}), and~(\ref{77o}), $R^{T2,\tau}_f({\bf y})$ is less than $(h+b)\cdot \sum_{t=\tau+2}^T\mathbb{E}_f[Z_t]$, where
\begin{equation}\begin{array}{l}
Z_t=\sum_{d=\hat y_t}^{\hat y_{t-\tau}-D_{t-\tau}-\cdots-D_{t-1}-1}\mid F_f(d)-\beta\mid\times\\
\;\;\;\;\;\;\;\;\;\times{\bf 1}(\beta\leq\hat F_{t-1}(\hat y_t)\leq \hat F_{t-1}(\hat y_{t-\tau}-D_{t-\tau}-\cdots-D_{t-1}-1)<\beta+\tau_T/(t-\tau)).\end{array}\end{equation}
In general,
\begin{equation}\label{gen}
Z_t\leq\bar d.
\end{equation}
Noting in turn that $\mid F_f(d)-\hat F_{t-1}(d)\mid\leq \delta_V(f,\hat f_{t-1})$, which is less than $\sqrt{\mbox{D}_{KL}(\hat f_{t-1}||f)/2}$ due to Pinsker's inequality~(\ref{pinsker}), we also have
\begin{equation}\label{impotent}\begin{array}{l}
Z_t\leq \sum_{d=\hat y_t}^{\hat y_{t-\tau}-D_{t-\tau}-\cdots-D_{t-1}-1}(\mid F_f(d)-\hat F_{t-1}(d)\mid+\mid\hat F_{t-1}(d)-\beta\mid)\times\\
\;\;\;\;\;\;\;\;\;\times{\bf 1}(\beta\leq\hat F_{t-1}(\hat y_t)\leq \hat F_{t-1}(\hat y_{t-\tau}-D_{t-\tau}-\cdots-D_{t-1}-1)<\beta+\tau_T/(t-\tau))\\
\;\;\;\; \leq \bar d\cdot [\sqrt{\mbox{D}_{KL}(\hat f_{t-1}||f)/2}+\tau_T/(t-\tau)].
\end{array}\end{equation}
So for a sequence $\varepsilon_t$,
\begin{equation}\begin{array}{l}
R^{T2,\tau}_f({\bf y})\leq (h+b)\cdot \sum_{t=\tau+2}^T\mathbb{E}_f[Z_t|\mbox{D}_{KL}(\hat f_{t-1}||f)\leq\varepsilon_t]\cdot\mathbb{P}_f[\mbox{D}_{KL}(\hat f_{t-1}||f)\leq\varepsilon_t]\\
\;\;\;\;\;\;\;\;\;\;\;\;\;\;\;\;\;\;+(h+b)\cdot\sum_{t=\tau+2}^T\mathbb{E}_f[Z_t|\mbox{D}_{KL}(\hat f_{t-1}||f)>\varepsilon_t]\cdot\mathbb{P}_f[\mbox{D}_{K}(\hat f_{t-1}||f)>\varepsilon_t],
\end{array}\end{equation}
which by~(\ref{sanov-son}),~(\ref{gen}), and~(\ref{impotent}), is less than
\begin{equation}\label{chicago}
(h\bar d+b\bar d)\cdot\sum_{t=1}^T[\sqrt{\frac{\varepsilon_t}{2}}+t^{\bar d+1}\cdot\exp(-\varepsilon_t\cdot (t-1))+\frac{\tau_T}{t}].
\end{equation}
The situation we face is very similar to~(\ref{abv}) except for the $\tau_T/t$-term. So as in Proposition~\ref{1-bound-hp}, for any $\epsilon>0$, there are constants $C'_\epsilon$, $D'_\epsilon$, and $E$ so that
\begin{equation}
R^{T2,\tau}_f({\bf y})\leq C'_\epsilon+D'_\epsilon\cdot T^{1/2+\epsilon/2}+E \tau_T\cdot \ln T.
\end{equation}
Note the $E$-term stems from the $\tau_T/t$-term in~(\ref{chicago}). In view of%~(\ref{import-bound}),
~(\ref{okkk0}) and~(\ref{kkko}),
\begin{equation}
R^{T2}_f({\bf y})\leq C'_\epsilon\tau_T+D'_\epsilon\tau_T\cdot T^{1/2+\epsilon/2}+E\tau_T^{\;2}\cdot \ln T+(T-2)\cdot (h\bar d+b\bar d)\cdot (\tau_T+1)^{\;\bar d}\cdot (1-\gamma)^{\tau_T-\bar d+1},
\end{equation}
when $\tau_T$ is above the $\theta_\gamma$ defined right after~(\ref{bazaj}). Otherwise, we have almost the same inequality, albeit with the last term replaced by $(T-2)\cdot(h\bar d+b\bar d)$. Choose $\tau_T$ appropriately, say $\tau_T=\lfloor T^{\epsilon/2}\rfloor$. Then, as long as $T$ is large enough, say greater than some $T^0_{\epsilon,\gamma}$, we can ensure that $\lfloor T^{\epsilon/2}\rfloor$ is above $\theta_\gamma$. Very importantly, just because $\gamma\in (0,1]$, we can make sure that the last term, regardless whether $\lfloor T^{\epsilon/2}\rfloor$ is below or above $\theta_\gamma$, is always bounded from above by a positive constant $F_{\epsilon,\gamma}$. Thus,
\begin{equation}
R^{T2}_f({\bf y})\leq C'_\epsilon\cdot T^{\epsilon/2}+D'_{\epsilon}\cdot T^{1/2+\epsilon}+E\cdot T^\epsilon\cdot\ln T+F_{\epsilon,\gamma}.
\end{equation}
However, as long as $T$ is large enough, the $T^{1/2+\epsilon}$-sized term will dominate all other terms. A constant term can certainly cover the case when $T$ is not that large. Therefore, positive constants $C_{\epsilon,\gamma}$ and $D_\epsilon$ exist for the intended inequality
\begin{equation}
R^{T2}_f({\bf y})\leq C_{\epsilon,\gamma}+D_{\epsilon}\cdot T^{1/2+\epsilon}.
\end{equation}
Since $C_{\epsilon,(1-\beta)/2}$ can be used for cases with $\gamma\geq (1-\beta)/2>0$, we can have the intended bound, namely,
\begin{equation}\label{intended}
R^{T2}_f({\bf y})\leq C_{\epsilon}+D_{\epsilon}\cdot T^{1/2+\epsilon},
\end{equation}
as long as $\gamma$ stays above $(1-\beta)/2$.

We now turn to the second case with $\gamma\in [0,(1-\beta)/2)$. From~(\ref{ee74}), $R^{T2}_f({\bf y})$ is equal to
\begin{equation}\label{loose-b}\begin{array}{l}
\sum_{t=3}^T\sum_{s=2}^{t-1}\mathbb{E}_f[Q_f(\hat y_s-D_s-\cdots-D_{t-1})-Q_f(\hat y_t)|L(t)=s]\cdot \mathbb{P}_f[L(t)=s]\\
\;\;\;\;\;\;\;\;\;\;\;\;\leq (h\bar d+b\bar d)\cdot\sum_{t=3}^T\max_{s=2}^{t-1}\mathbb{P}_f[\hat y_s-D_s-\cdots-D_{t-1}\geq 1],
\end{array}\end{equation}
in which the inequality is attributable to the facts that $\sum_{s=2}^{t-1}\mathbb{P}_f[L(t)=s]\leq 1$ and that $Q_f(z)-Q_f(y)\leq (h\bar d+b\bar d)\cdot {\bf 1}(z\geq y+1)$ for $z\geq y$, which is easy to see from~(\ref{def0}). But
\begin{equation}\label{loose-c}
\mathbb{P}_f[\hat y_s-D_s-\cdots-D_{t-1}\geq 1]\leq  \mathbb{P}_f[\hat y_s\geq 1]\wedge (\sum_{d=1}^{\bar d}\mathbb{P}_f[\hat y_s\geq d]\cdot \mathbb{P}_f[D_s+\cdots+D_{t-1}\leq d-1]).
\end{equation}
Meanwhile, by~(\ref{newsboy}) and the current range of $\gamma$,
\begin{equation}
\mathbb{P}_f[\hat y_s\geq 1]=\mathbb{P}_f[\hat F_{s-1}(0)<\beta]\leq \mathbb{P}_f[\delta_V(f,\hat f_{s-1})>1-\beta-\gamma],
\end{equation}
which, due to Pinsker's inequality~(\ref{pinsker}), is below $\mathbb{P}_f[\mbox{D}_{K}(\hat f_{s-1}||f)>2\ \cdot(1-\beta-\gamma)^2]$. But by Sanov's~(\ref{ssnv}), the latter is below $s^{\bar d+1}\cdot\exp(-2\cdot(1-\beta-\gamma)^2\cdot(s-1))$. Thus,
\begin{equation}\label{a22}
\mathbb{P}_f[\hat y_s\geq 1]\leq s^{\bar d+1}\cdot\exp(-2\cdot(1-\beta-\gamma)^2\cdot(s-1)).
\end{equation}
Now for $d=1,2,...,\bar d$, let $\gamma_d=1-F_f(d-1)$. Our setup is such that $0\leq \gamma_{\bar d}\leq\gamma_{\bar d-1}\leq\cdots\leq \gamma_1=\gamma<(1-\beta)/2$. %By~(\ref{newsboy}), we have $\hat y_s\geq d$ if and only if $\hat F_{s-1}(d-1)<\beta$.
Again due to~(\ref{newsboy}),
\begin{equation}\label{lucky}
\mathbb{P}_f[\hat  y_s\geq d]=\mathbb{P}_f[\hat F_{s-1}(d-1)<\beta]=\mathbb{P}_f[\sum_{\tau=1}^{s-1}{\bf 1}(D_\tau\geq d)>(1-\beta)\cdot (s-1)].
\end{equation}
Note that ${\bf 1}(D_1\geq d),{\bf 1}(D_2\geq d),...,{\bf 1}(D_{s-1}\geq  d)$ are independent Bernoulli random variables with mean $\gamma_d$, and hence $\sum_{\tau=1}^{s-1}{\bf 1}(D_\tau\geq d)$ is a Binomial random variable with mean $\gamma_d\cdot(s-1)$. So by Markov's inequality, the rightmost term in~(\ref{lucky}) is below
\begin{equation}
\frac{\mathbb{E}_f[\sum_{\tau=1}^{s-1}{\bf 1}(D_\tau\geq d)]}{(1-\beta)\cdot(s-1)}=\frac{\gamma_d}{1-\beta}.
\end{equation}
Therefore,
\begin{equation}\label{okko}
\mathbb{P}_f[\hat y_s\geq d]\leq \frac{\gamma_d}{1-\beta}.
\end{equation}
Also, it is easy to see that
\begin{equation}\label{a330}
\mathbb{P}_f[D_s+\cdots+D_{t-1}\leq d-1]\leq (1-\gamma_d)^{t-s}.
\end{equation}

Combining~(\ref{loose-c}),~(\ref{a22}),~(\ref{okko}), and~(\ref{a330}), we obtain
\begin{equation}\label{joye}\begin{array}{l}
\max_{s=2}^{t-1}\mathbb{P}_f[\hat y_s-D_s-\cdots-D_{t-1}\geq 1]\\
\;\;\;\leq \max_{s=2}^{t-1}[s^{\bar d+1}\cdot \exp(-2\cdot(1-\beta-\gamma)^2\cdot(s-1))]\wedge\\
 \;\;\;\;\;\;\;\;\;\;\;\;\;\;\;\;\;\;\;\;\;\wedge[\sum_{d=1}^{\bar d}(\gamma_d/(1-\beta))\cdot (1-\gamma_d)^{t-s}].
\end{array}\end{equation}
Consider $a(\gamma,\tau)\equiv\tau^{\bar d+1}\cdot \exp(-2\cdot(1-\beta-\gamma)^2\cdot(\tau-1))$. There exists a $t_0\geq 1$ so that for any $t\geq t_0$, the function $a((1-\beta)/2,t)$ will decrease with $t$, and also
\begin{equation}\label{azaza}
a(\frac{1-\beta}{2},t)=t^{\bar d+1}\cdot\exp(-\frac{(1-\beta)^2}{2}\cdot(t-1))<\exp(-\frac{(1-\beta)^2\cdot t}{4}).
\end{equation}
Note also that $a(\gamma,s)<a((1-\beta)/2,s)$ for $\gamma\in (0,(1-\beta)/2)$.
%For $s\geq t/2\geq t_0$, consider $a_\gamma(s)$. Then by~(\ref{azaza}), the term is less than $\exp(-(1-\beta)^2s/4)$.
Next, consider $b(\gamma',\tau)\equiv \gamma'\cdot (1-\gamma')^\tau$. Note that
\begin{equation}
\frac{\partial b(\gamma',\tau)}{\partial\gamma'}=(1-\gamma')^{\tau-1}\cdot [1-(\tau+1)\cdot\gamma'],
\end{equation}
and
\begin{equation}
\frac{\partial^2 b(\gamma',\tau)}{\partial(\gamma')^2}=-\tau\cdot (1-\gamma')^{\tau-2}\cdot [2-(\tau+1)\cdot\gamma'].
\end{equation}
So the $b$-maximizing $\gamma'$ is $\gamma^*_\tau=1/(\tau+1)$. Plugging back, we have
\begin{equation}\label{bzaza}
b(\gamma^*_\tau,\tau)=\frac{1}{\tau+1}\cdot (1-\frac{1}{\tau+1})^{\tau}=\frac{1}{\tau+1}\cdot\frac{1}{(1+1/\tau)^\tau}.
\end{equation}
Note that $\lim_{\tau\rightarrow +\infty}(1+1/\tau)^\tau=e$, the natural logarithmic base which is above 2. So when $\tau$ is large enough, say greater than some $t_1$, the above will be below $1/(2\tau+2)$.

For $T\geq 2t_0+2t_1$, the upper bound in~(\ref{joye}) is further bounded by a constant plus
\begin{equation}
\sum_{t=2t_0+2t_1+1}^T(\max_{s=2}^{\lfloor t/2\rfloor}\sum_{d=1}^{\bar d}\frac{\gamma_d}{1-\beta}\cdot (1-\gamma_d)^{t-s}+\max_{s=\lfloor t/2\rfloor+1}^{t-1}s^{\bar d+1}\cdot\exp(-2\cdot(1-\beta-\gamma)^2\cdot(s-1)),
\end{equation}
which, according to the above from~(\ref{azaza}) to~(\ref{bzaza}), is below
\begin{equation}
\sum_{t=2t_0+2t_1+1}^T[\frac{\bar d}{1-\beta}\cdot\max_{s=2}^{\lfloor t/2\rfloor}\frac{1}{2t-2s+2}+\max_{s=\lfloor t/2\rfloor+1}^{t-1}\exp(-\frac{(1-\beta)^2\cdot s}{4})].
\end{equation}
But this is smaller than
\begin{equation}
\sum_{t=2t_0+2t_1+1}^T[\frac{\bar d}{(1-\beta)\cdot t}+\exp(-\frac{(1-\beta)^2\cdot t}{8})],
\end{equation}
which has a constant-plus-$\ln T$ bound. So, there exist positive constants $E$ and $F$ so that
\begin{equation}\label{tendin}
R^{T2}_f({\bf y})\leq E+F\cdot \ln T,
\end{equation}
for any $\gamma\in (0,(1-\beta)/2)$. Now between~(\ref{intended}) and~(\ref{tendin}), only the former has to be used when $C_\epsilon$ is made large enough. We therefore have the intended bound. \qed

\end{document}